\documentclass[conference]{IEEEtran}
\IEEEoverridecommandlockouts
\usepackage{cite}
\usepackage{amsmath,amssymb,amsfonts}
\usepackage[utf8]{inputenc}
\usepackage{algorithmic}
\DeclareUnicodeCharacter{05F3}{'}

\usepackage{graphicx}
\usepackage{tabularx} 
\usepackage{textcomp}
\usepackage{xcolor}
\def\BibTeX{{\rm B\kern-.05em{\sc i\kern-.025em b}\kern-.08em
    T\kern-.1667em\lower.7ex\hbox{E}\kern-.125emX}}
\begin{document}

\title{High Order Reasoning for Time Critical Recommendation in Evidence-based Medicine \\

}

\author{\IEEEauthorblockN{1\textsuperscript{st} Manjiang Yu}
\IEEEauthorblockA{\textit{School of Electrical Engineering and Computer Science} \\
\textit{University of Queensland}\\
Brisbane, Australia \\
manjiang.yu@uqconnect.edu.au}
\and
\IEEEauthorblockN{2\textsuperscript{nd} Xueli}
\IEEEauthorblockA{\textit{School of Electrical Engineering and Computer Science} \\
\textit{University of Queensland}\\
Brisbane, Australia \\
xueli@eecs.uq.edu.au}

}

\maketitle

\begin{abstract}

In time-critical decisions, human decision-makers can interact with AI-enabled situation-aware software to evaluate many imminent and possible scenarios, retrieve billions of facts, and estimate different outcomes based on trillions of parameters in a fraction of a second. In high-order reasoning, "what-if" questions can be used to challenge the assumptions or pre-conditions of the reasoning, "why-not" questions can be used to challenge on the method applied in the reasoning, "so-what" questions can be used to challenge the purpose of the decision, and "how-about" questions can be used to challenge the applicability of the method. When above high-order reasoning questions are applied to assist human decision-making, it can help humans to make time-critical decisions and avoid false-negative or false-positive types of errors. In this paper, we present a model of high-order reasoning to offer recommendations in evidence-based medicine in a time-critical fashion for the applications in intensive care unit. The Large Language Model (LLM) is used in our system. The experiments demonstrated after system message and few-shot learning, the LLM exhibited optimal performance in the "What-if" high-order reasoning scenario, achieving a similarity of 88.52\% with the treatment plans of human doctors. In the "Why-not" scenario, the best-performing model tended to opt for alternative treatment plans in 70\% of cases for patients who died after being discharged from the ICU. In the "So-what" scenario, the optimal model provided a detailed analysis of the motivation and significance of treatment plans for ICU patients, with its reasoning achieving a similarity of 55.6\% with actual diagnostic information. In the "How-about" scenario, the top-performing LLM demonstrated a content similarity of 66.5\% in designing treatment plans transferring for similar diseases. Meanwhile, LLMs managed to predict the life status of patients after their discharge from the ICU with an accuracy of 70\%.

\end{abstract}


\section{Introduction}
People can think fast and slow\cite{daniel2017thinking }; in the Intensive Care Unit (ICU), it's crucial for doctors to think slowly to ensure the accuracy of each operation on patients. However, efficiency is equally important in these high-stakes environments. Large language models (LLMs), with their ability to rapidly process vast volumes of data, high-speed computing power, and capacity to coordinate with numerous agents, emerge as potential tools. They can augment the decision-making process in the ICU, providing timely, data-driven insights that support the precision and efficiency required in critical care. In the demanding Intensive Care Unit (ICU) environment, where resources are scarce and medical staff face time pressure and heavy workloads, there's a pressing need for strategies to alleviate these burdens. The uneven global distribution of ICU resources further highlights this need. Reducing decision-making pressure could greatly improve ICU resource scalability and patient outcomes.The emergence of Large Language Models like GPT has opened new possibilities in various sectors, including healthcare. These models are increasingly recognized for their ability to enhance medical efficiency, having shown proficiency in simple medical diagnostics and efficient management of electronic medical records. However, these tasks are less complex compared to the intricate decision-making required in the ICU.This thesis explores whether LLMs can be applied to complex medical decision-making in the ICU, involving high-order reasoning. If successful, this could significantly impact AI-assisted treatment and medical education in healthcare.
High-order reasoning involves critical, analytical, and creative thinking, essential in ICUs where complex patient care is required\cite{gopalan2019decision}. It surpasses basic cognitive skills, enabling healthcare professionals to make timely, precise decisions and develop personalized treatment plans amidst diverse patient conditions. With advancements in AI and NLP, the role of technology in augmenting high-order reasoning is gaining attention. AI and NLP-powered chatbots can analyze extensive data and interpret complex medical scenarios, becoming valuable assistants in clinicians' decision-making processes\cite{hacker2023understanding}.

\subsection{High order Reasoning with LLM for ICU treatment recommendations}

The rise of Large Language Models (LLMs) in natural language processing and AI has significant implications for healthcare, especially in the high-pressure setting of Intensive Care Units (ICUs). These advanced models present opportunities for smart, swift, and precise treatment recommendations, meeting the critical demand for efficient decision-making tools in ICUs, where time and resources are often limited. Nonetheless, incorporating these models into ICU workflows presents certain challenges, particularly in evaluating their capability to perform high-order reasoning in critical care. Specific problems include:
\begin{enumerate}
    \item How effective are LLMs at understanding complex medical data in intricate scenarios and providing accurate and reliable treatment advice in real-time? 
    
    \item How can we ensure the content generated by LLMs is harmless and accurate?

     \item How do the suggestions provided by LLMs compare in accuracy, reliability, and applicability to those offered by experienced ICU doctors? 

      \item What strategies can be adopted to enhance the higher-order reasoning abilities of LLMs in the context of ICU treatment advice? 
\end{enumerate}

This study evaluates the effectiveness of Large Language Models in providing treatment advice in ICUs, using real ICU data.  It investigates techniques such as prompt engineering and few-shot learning to enhance the sophisticated reasoning capabilities of these models in a complex, fast-paced setting. The research concentrates on the hurdles of incorporating these AI tools into ICU workflows, with a particular focus on ensuring their advice's accuracy, reliability, and safety in comparison to seasoned physicians. The objective is to thoroughly evaluate and improve the application of these models in ICU treatment decision-making. Specific goals include:

\begin{enumerate}
    \item \textbf{Present and evaluate LLMs' High-Order reasoning capabilities in the ICU Environment:} Using eICU to simulate real ICU scenarios, and employing system message, prompt engineering, and few-shot learning to promote LLM's demonstration of its high-order reasoning capabilities. Assess the accuracy, reliability, and applicability of LLMs in providing higher-order reasoning, using complex, real-world ICU data. Compare the LLM-generated suggestions to the advice of experienced ICU doctors to measure the models’ capability and reliability.
    
    \item \textbf{Exploring the impact of various technologies on LLM's high-order reasoning in the medical field:} Explore the impact of technologies like zero-shot learning and fine-tuning on LLMs’ reasoning capacities. compare the performance of LLMs with different techniques in complex high-order reasoning medical decision scenarios.
\end{enumerate}

Each objective is meticulously designed to foster a comprehensive understanding and tangible enhancement of LLMs' applications in ICU treatment advice.

\begin{figure}[htbp]
\centerline{\includegraphics[width=0.5\textwidth]{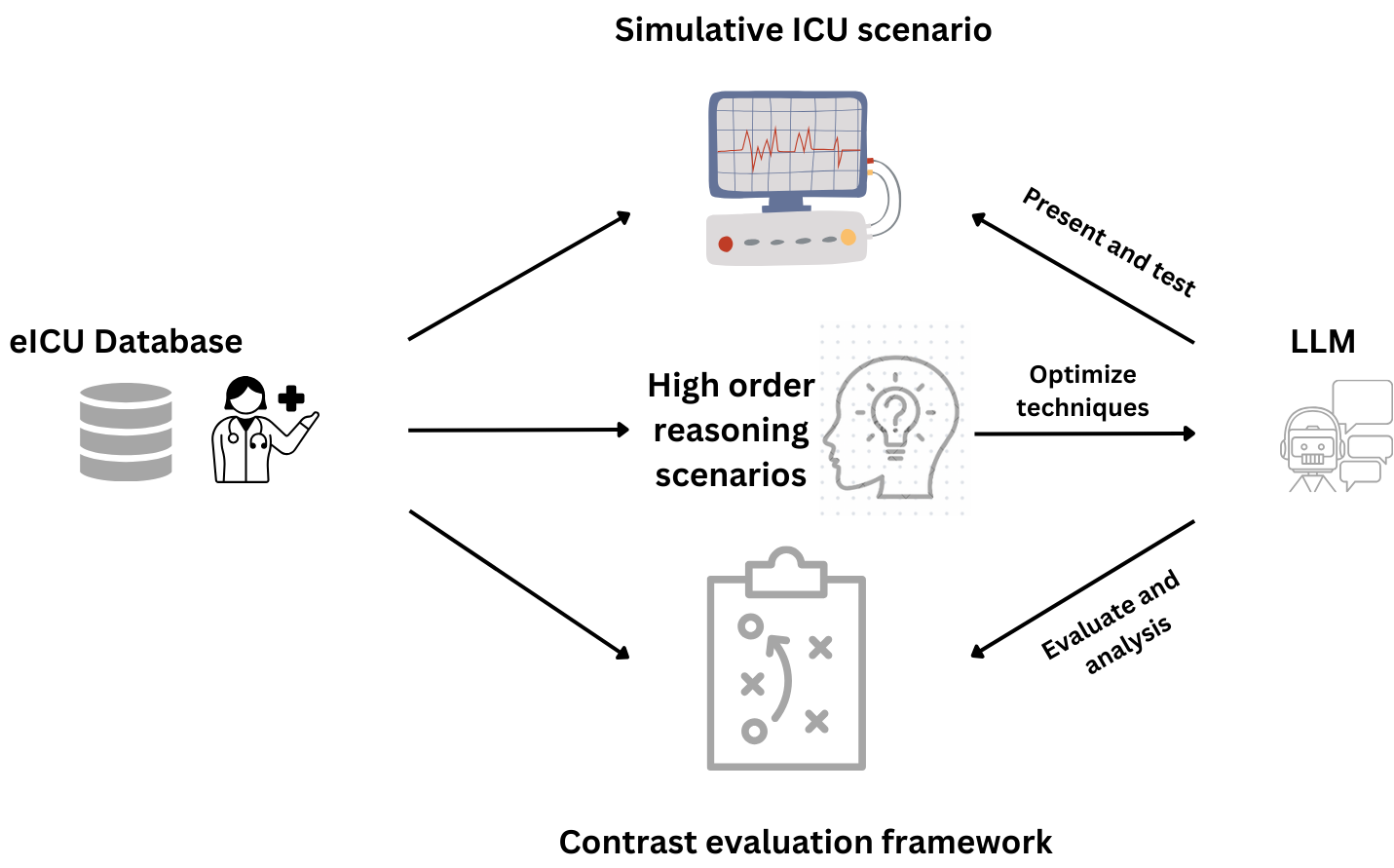}}
\caption{Intuition diagram}
\label{fig}
\end{figure}

\subsection{Contribution}

The study introduces four scenarios for assessing high-order reasoning in healthcare, specifically in ICUs, by employing technologies like system messages and few-shot learning to evaluate the reasoning abilities of LLMs in medical environments. Using the eICU database and a new 'contrast evaluation framework', it measures LLMs against physician decisions, innovating in the validation of LLM capabilities in critical care. The research also examines the impact of fine-tuning and zero-shot learning on LLM performance and compares different models' abilities to predict hospital discharge outcomes, contributing valuable insights into LLMs' potential in healthcare.
The study’s focus is not immediate clinical application but to inform medical education, helping students and professionals better understand and navigate complex medical situations through LLM-supported simulations. This approach aims to enrich medical training by integrating advanced technologies with conventional teaching methods, preparing future healthcare.

\section{Related work}

    \par

   \subsection{AI and NLP in Healthcare}
Artificial intelligence (AI), including machine learning, natural language processing (NLP), and deep learning, has significantly advanced healthcare in the past decade. These technologies have been applied in diagnosing various diseases such as stroke, Alzheimer's disease, skin cancer, and neurological disorders. The combination of AI, NLP, and Large Language Models (LLMs) has shown great promise in healthcare, particularly in cases of acute ischemic stroke and other conditions.

For instance, Khedher et al.~\cite{khedher2015early} used support vector machines for the early detection of Alzheimer's disease. Fiszman et al.~\cite{fiszman2000automatic} initially proposed using NLP for medical diagnosis, applying it to detect acute bacterial pneumonia from chest X-ray reports. Naveed et al.~\cite{afzal2017mining} achieved a 90\% classification accuracy in mining peripheral arterial disease cases from clinical narratives. Qiuyue Zhong et al.~\cite{zhong2018screening} used NLP to screen for suicidal behavior in pregnant women at the Mayo Clinic, which also offers NLP-as-a-service for various healthcare applications~\cite{wen2019desiderata}. 

During the COVID-19 pandemic, AI-powered chatbots, as used by Ay Carriere's team~\cite{carriere2021case}, helped address patient needs and reduce healthcare professional burdens. Tvardik et al.~\cite{Tvardik2018} employed NLP to detect healthcare-associated infections (HAI) with high precision. Wee et al.~\cite{Wee2022} explored machine learning for classifying medical referrals, enhancing efficiency and resource allocation. Tsoukalas et al.~\cite{Tsoukalas2015} developed a machine learning-based decision support system for patients with sepsis, demonstrating its potential in improving patient outcomes.

These studies collectively highlight the transformative role of AI and NLP in various aspects of healthcare.

    \par

    \subsubsection{LLMs in Healthcare}
    
    In recent years, large language modeling (LLM) has been transforming healthcare, with applications ranging from diagnosing diseases to developing treatment plans. \cite{vaswani2017attention} Key developments include the Generative Pre-Training Transformer (GPT) models by OpenAI, such as GPT-3 and its conversational variant, ChatGPT. \cite{floridi2020gpt} Marco et al. explored ChatGPT's feasibility in healthcare, particularly in understanding and summarizing clinical scenarios. \cite{gilson2022well} ChatGPT also demonstrated potential in passing the USMLE, a comprehensive medical exam. \cite{makhoul2020objective} Kung et al. evaluated ChatGPT on the USMLE, finding it achieved passing scores across all three steps. \cite{kung2023performance} Victor Tseng's study further confirmed ChatGPT's impressive performance against non-medical professionals. \cite{brown2023chatgpt} GPT-4, a more advanced LLM, exhibits human-level performance on various benchmarks, including the USMLE. \cite{openai2023gpt4} Researchers from OpenAI and Microsoft found GPT-4 outperformed earlier models and specialized medical models like Med-PaLM. \cite{nori2023capabilities} Zhang et al. assessed GPT-4's proficiency in disease classification using a real-world health record database, noting its high accuracy but also some limitations. \cite{zhang2023potential} Thirunavukarasu et al. explored the technical limitations and barriers of LLMs in healthcare, emphasizing the need for further validation. \cite{thirunavukarasu2023large} Google's research team investigated LLMs in encoding clinical knowledge, introducing MultiMedQA and HealthSearchQA for evaluating models like PaLM and Flan-PaLM. \cite{singhal2022large} Despite Flan-PaLM's state-of-the-art accuracy, human assessments revealed gaps, leading to the development of Med-PaLM, which showed promise but still requires further development for clinical applications. This comprehensive overview highlights the evolving role of LLMs in healthcare, underscoring their potential and the need for ongoing research and development.
\subsubsection{Technoqiues in Optimizing}

In the field of large language models, advancements in specific technologies are pivotal for their efficacy in specialized domains like healthcare. This paper delves into key strategies such as system messaging, prompt engineering, few-shot learning, zero-shot learning, and fine-tuning, highlighting their roles and potential applications. System messaging involves guiding the model's responses through prompts, playing a vital role in directing LLM behavior and breaking domain-specific limitations \cite{kong2023better}. Prompt engineering, a crucial skill for effective interaction, involves crafting well-structured prompts to elicit specific responses, ensuring contextually appropriate insights \cite{liu2023pre, white2023prompt}.  Few-shot learning, as demonstrated by models like GPT-3, allows learning from limited data, illustrating adaptability and pattern application in new situations \cite{perez2021true, wang2020generalizing}. Zero-shot learning enables handling tasks without prior examples, showcasing utility in dynamic settings \cite{wei2021finetuned, kojima2022large}. Fine-tuning adjusts pre-trained models for specific tasks, improving accuracy and relevance in particular fields \cite{wei2021finetuned, kojima2022large}.  These technologies, particularly relevant in healthcare, offer solutions for complex decision-making and reasoning. The subsequent chapters will explore the integration of these techniques in higher-order reasoning for ICU treatment recommendations, providing a comprehensive understanding of their practical application and impact in critical care settings.

    \par
    \subsection{Chatbots and Decision Support Systems in Medicine}

    The integration of artificial intelligence in digital healthcare has led to the emergence of chatbots and decision support systems (DSS) as transformative tools. These digital solutions facilitate interactions between healthcare professionals and patients, aiding in decision-making, symptom checking, and providing medical information. They promise to enhance healthcare delivery's quality and efficiency through accurate clinical decisions, supported by robust data and algorithms. However, challenges like data privacy, the need for continuous updates, and algorithmic bias remain. Flora Amato et al. proposed a chatbot system, HOLMes, to improve healthcare efficiency and quality \cite{amato2017chatbots}. I-CHING and Jiun-De Yu developed a machine learning-based medical chatbot, MLCF, drawing knowledge from open data sources, though its effectiveness is limited by training set size and quality \cite{hsu2022medical}. Pin Ni's team created a chatbot using knowledge graphs and deep learning for medical advice, enhancing understanding and precision \cite{ni2022knowledge}. Ann Sato's team developed a chatbot for preliminary genetic screening for HBOC, employing Watson's functionality and NLP techniques \cite{Sato2021}. Bao, Qiming et al. designed the HHH, an online medical chatbot system using knowledge graphs and hierarchical bidirectional attention, although it has limitations in addressing diverse medical issues and understanding context \cite{Bao2020}.

    \par
    \subsection{High-order Reasoning}

     High-order reasoning, a cognitive process for solving complex problems or making decisions, is rooted in educational taxonomies like Bloom's Taxonomy. This taxonomy outlines several levels of higher-order thinking: remembering basic facts \cite{lewis1993defining}, understanding main ideas and concepts, applying concepts in real-world situations, analyzing topics from multiple perspectives, synthesizing conclusions, evaluating arguments, and creating new patterns or structures \cite{british_council}. Lewis and Smith.D simplify this as interrelating new and stored information to achieve a purpose or solve complex situations \cite{adams2015bloom}. Lindsey Engle Richard and Nina Simms emphasize the role of analogical cognitive mechanisms in higher-order thinking, focusing on relationships rather than discrete phenomena \cite{richland2015analogy}. In artificial intelligence, higher-order reasoning involves advanced cognitive processes like reasoning, decision-making, and problem-solving \cite{sinha2021chatgpt}. This goes beyond simple data analysis, requiring the understanding and manipulation of abstract concepts and contextual relevance. Alexander Muacevic and John R Adler demonstrate ChatGpt's capabilities in higher-order reasoning in pathology, evaluating its speed, reliability, and accuracy in answering high-order questions \cite{sinha2021chatgpt}.

    \subsection{Research gap}

     Despite significant interest in AI's role in healthcare, particularly in ICUs, notable research gaps remain:

\begin{itemize}
    \item \textbf{Differing complexity in application scenarios:}  Current research in medical AI primarily addresses basic medical diagnostics and electronic health record processing. However, these applications differ significantly from the complex high-order reasoning and decision-making needed in Intensive Care Units (ICUs). The ICU environment presents challenges in terms of data volume, variety, and the critical nature of decisions. This complexity makes research in this area both theoretically significant and practically urgent. Such studies expand our understanding of high-order reasoning in medical decision-making and establish a foundation for future applications in more complex medical scenarios.
    
    \item \textbf{Incompleteness of the Evaluation Framework: }  The existing evaluation frameworks for assessing the performance of Large Language Models in the healthcare sector are scarce and often lack objectivity and completeness. Many of the current approaches depend on manual assessments by human clinicians or simply evaluate based on the accuracy of multiple-choice questions. Neither of these methods is suitable or comprehensive enough to gauge the high-order reasoning capabilities in the ICU context as proposed by our study. Consequently, we have introduced a novel assessment paradigm known as the contrast evaluation framework, based on eICU databases. This framework allows us to evaluate the high-order reasoning abilities in the ICU setting in a thorough and unbiased manner. Furthermore, it enables us to assess the reliability and accuracy of the language model's output to a considerable extent. This innovative framework is a step towards a more rigorous and complete assessment mechanism, designed to meet the nuanced needs of high-stakes medical decision-making.
    
    \item \textbf{Evaluating the Impact of Techniques on LLM's High-Order Reasoning in Healthcare: }A multitude of techniques is available to enhance the learning process of Large Language Models for downstream tasks, significantly influencing their performance across diverse applications. Techniques such as system messages, few-shot learning, zero-shot learning, and fine-tuning hold the potential to refine model's capabilities, yet their utility in improving the performance specifically in the healthcare domain remains underexploited. In this study, we not only delve into the influence of various techniques on the high-order reasoning abilities of LLMs within the medical field but also draw comparisons between different LLMs utilizing distinct learning approaches. Our comprehensive analysis sheds light on their respective efficacies in executing advanced reasoning tasks. This research serves as a valuable benchmark and guide for future scholars, providing insights and tools to propel the evolution of LLMs in healthcare and beyond.
    
\end{itemize}

Given these gaps, there is an urgent need for a comprehensive investigation into the higher-order reasoning capabilities of chatbots, especially for treatment recommendations in ICUs. Exploration of this area not only offers the prospect of improving the quality of patient care but also contributes to the broader narrative of AI-driven medical education in healthcare.

\section{Dataset}

   This study utilizes the eICU dataset \cite{pollard2018eicu}, a comprehensive ICU database from the US, providing de-identified, high-resolution data on ICU admissions. Access to this public database requires registration, human research training, and a data use agreement. The eICU's structured documentation and diagnosis-treatment correlation led to its selection over the MIMIC-III database \cite{johnson2016mimic}. Its schema includes a non-normalized structure where tables are linked via unique identifiers like PatientUnitStayId, with each patient represented by a uniquePid and hospitalizations by PatientHealthSystemStayId, ensuring efficient data management for research and machine learning applications.

\begin{figure}[htbp]
\centerline{\includegraphics[width=0.5\textwidth]{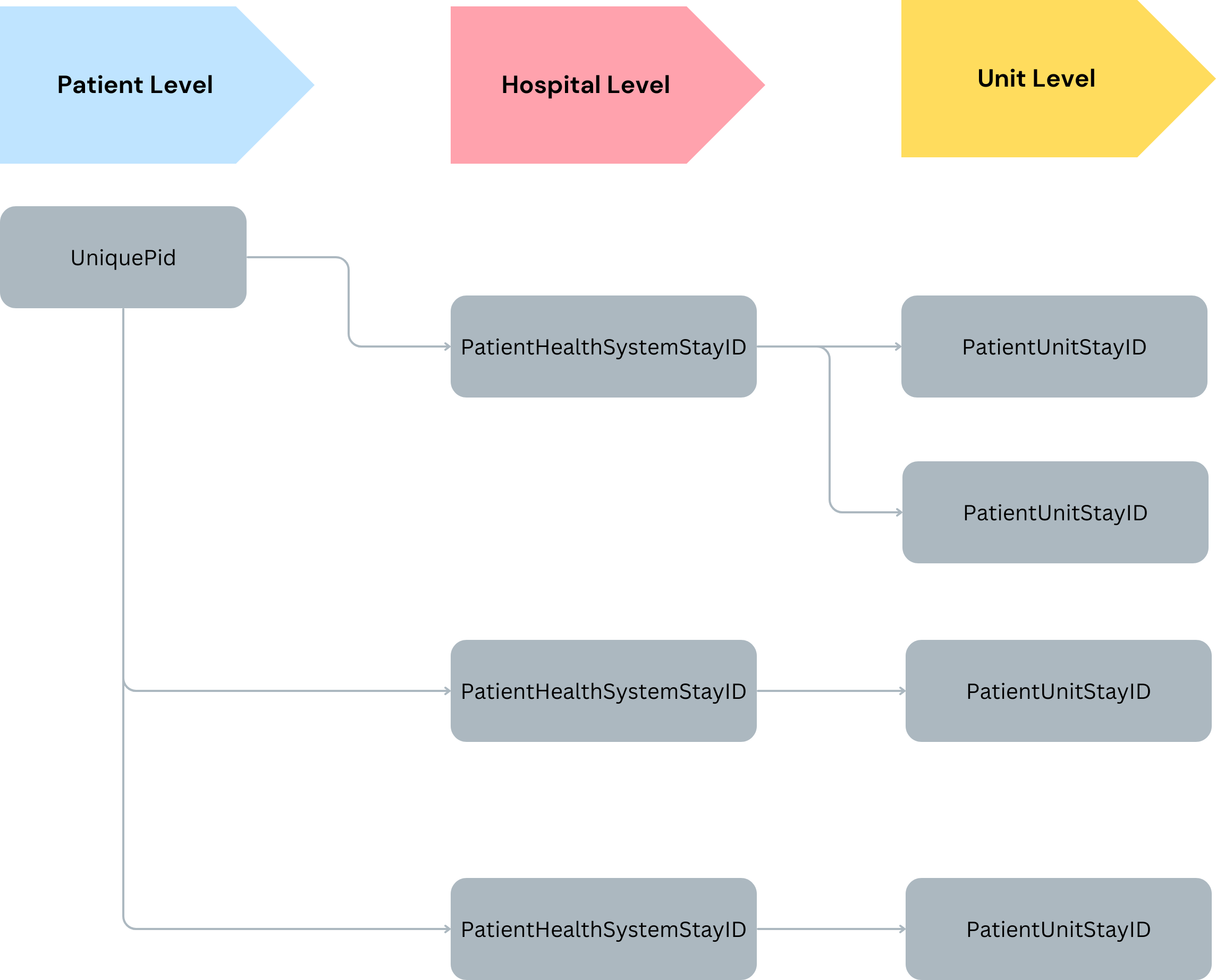}}
\caption{Organization of patient tracking information}
\label{fig}
\end{figure}

The first table in the database of the eICU used in this study is the patient table. \cite{pollard2018eicu}Records of Patient Information are Stored in the Patient Table Records of patient information are stored in the Patient table, which contains the three key identifiers mentioned earlier (PatientUnitStayId, PatientHealthSystemStayId, and uniquePid). The Patient table details administrative information about the patient, including time of admission and discharge, unit type, source of admission, place of discharge, and vital sign data at the time of the patient's discharge. In addition, the patient table summarizes demographic data about the patient, such as age (age $>$ 89 years is labeled "$>$ 89"), race, height, and weight, among other key information. The table contains the three key identifiers mentioned earlier (PatientUnitStayId, PatientHealthSystemStayId, uniquePid).

this table record all diagnosis in the duration of ICU period of a patient. Each diagnosis has its own time mark which is “Offset” column records the Accumulated time after the patient was admitted entering the ICU.

this table records all treatment in the duration of ICU period of a patient. Each treatment has its own time mark which is Offset” column records the Accumulated time after the patient was admitted entering the ICU.

This table records the vital signs data in the duration of the ICU of a patient. The records are periodic; each record has its own time mark, which is the “Offset” column, which records the accumulated time after the patient was admitted entering the ICU. \cite{pollard2018eicu} Continuously measured vital sign data were stored in the life cycle chart; these vital signs included heart rate, respiratory rate, oxygen saturation, body temperature, invasive arterial blood pressure, pulmonary artery pressure, ST-segment levels, and intracranial pressure (ICP). Initially, these vital sign data were collected at 1-minute intervals, and the median number over a 5-minute period was archived in the eICU-CRD database.

\section{Methodology}

In order to systematically explore and demonstrate the potential of large language models for higher-order reasoning in complex healthcare scenarios, we employ a series of innovative methodologies. Firstly, we design four \cite{li2023artificial} high-order reasoning contexts: what-if, why-not, so-what, and how-about. This advanced reasoning scenario framework serves as a foundation for demonstrating the higher-order reasoning capabilities of large language models, with each scenario being applicable to various complex medical situations in the ICU. The higher-order reasoning ability of large language models is also flexibly exhibited within this framework. In addition, based on the eICU dataset, we designed specific evaluation criteria for these four high-order reasoning scenarios. The eICU dataset provides valuable data support for our study due to its rich medical information and real-world application scenarios. Through this real-world data-based evaluation, we were able to objectively demonstrate the higher-order reasoning capability.

\begin{figure}[htbp]
\centerline{\includegraphics[width=0.5\textwidth]{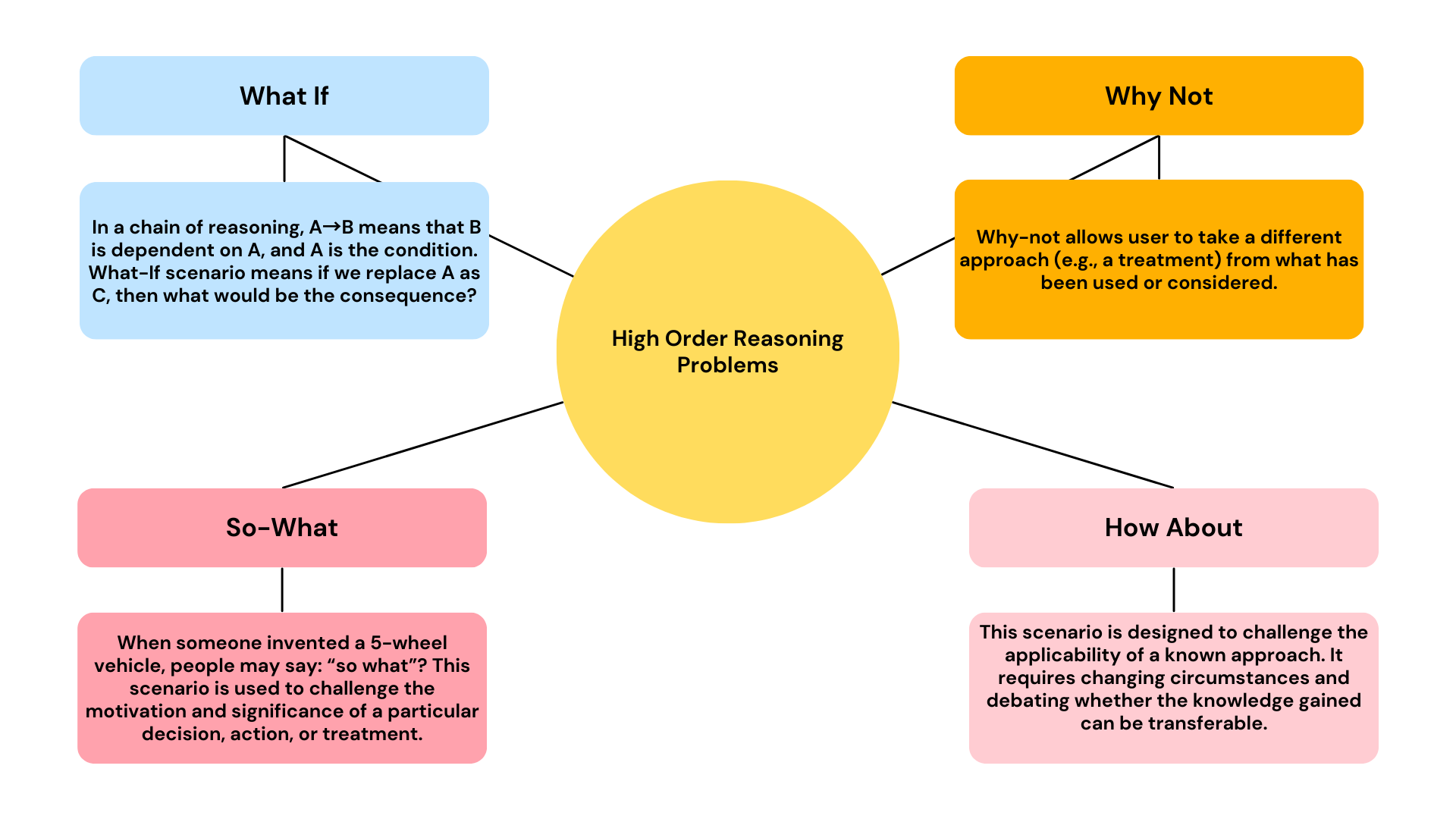}}
\caption{High order reasoning scenarios}
\label{fig}
\end{figure}

Second, to further challenge and test the comprehensive reasoning capability, we designed a complex discharge status prediction task. This task involves a large amount of time-series medical information and requires the model to speculate on the survival status of a patient at the time of hospital discharge, given this information. This is not only an important task to test model's performance in real-world applications, but also another core part of our implementation of higher-order inference methods.
\subsection{Predefinition of high-order reasoning (system message, prompt engineering, and few-shot learning)}Predefining high-order reasoning in Large Language Models involves setting roles and outputs to guide their reasoning capabilities. This setup involves creating predefined scenarios, utilizing system messages, and applying few-shot learning techniques. System messages play a key role in optimizing model performance while addressing inherent limitations. For instance, specific adaptations in models like ChatGPT, a variation of the GPT series, include safeguards (like Instruct GPT) against generating biased or harmful content. As a result, ChatGPT refrains from delivering sensitive information, such as instructions for making a bomb. However, these safeguards can sometimes restrict the model's full potential. In healthcare, ChatGPT's cautious response style yields general advice, not tapping into its full capabilities. System messages act as unseen directives for chatbots, steering their responses to meet set standards. Prompt engineering shapes LLM output to ensure alignment with defined goals. Illustrative scenarios for predefining higher-order reasoning aid in few-shot learning, where LLMs learn to tackle tasks effectively with limited examples, adapting similar reasoning patterns to meet set reasoning objectives \cite{brown2020language}.

\subsection{Contrast evaluation framework}

In addressing the need for a robust evaluation system for large language models (LLMs) in healthcare, current approaches often fall short, especially in complex medical scenarios. Existing frameworks typically hinge on simple medical diagnostics or rely on scoring systems by human doctors, which do not adequately capture the nuances of high-order reasoning in intricate medical contexts. To bridge this gap, our study introduces an innovative evaluation framework specifically designed for assessing the high-order reasoning capabilities of LLMs in the medical field, termed the contrast evaluation framework. Central to this framework is its foundation on a real ICU medical record database.

Within this framework, we construct distinct convergence situations for various high-order reasoning scenarios, based on the database. Each scenario is evaluated independently, ensuring a comprehensive and objective assessment. This contrasts with other current frameworks, which often struggle to accurately and rationally validate the content produced by LLMs in medical contexts. Our framework addresses this challenge effectively.

Next, we will delve into the four specific high-order reasoning scenarios and the task of predicting final life status. This thorough examination of LLMs' high-order reasoning abilities and the evaluation process offers a complete understanding of their potential in ICU settings. This approach not only highlights the capabilities of LLMs but also sets a new standard for their evaluation in the critical domain of healthcare.

\subsection{ What-if scenario}

\begin{figure}[htbp]
\centerline{\includegraphics[width=0.45\textwidth]{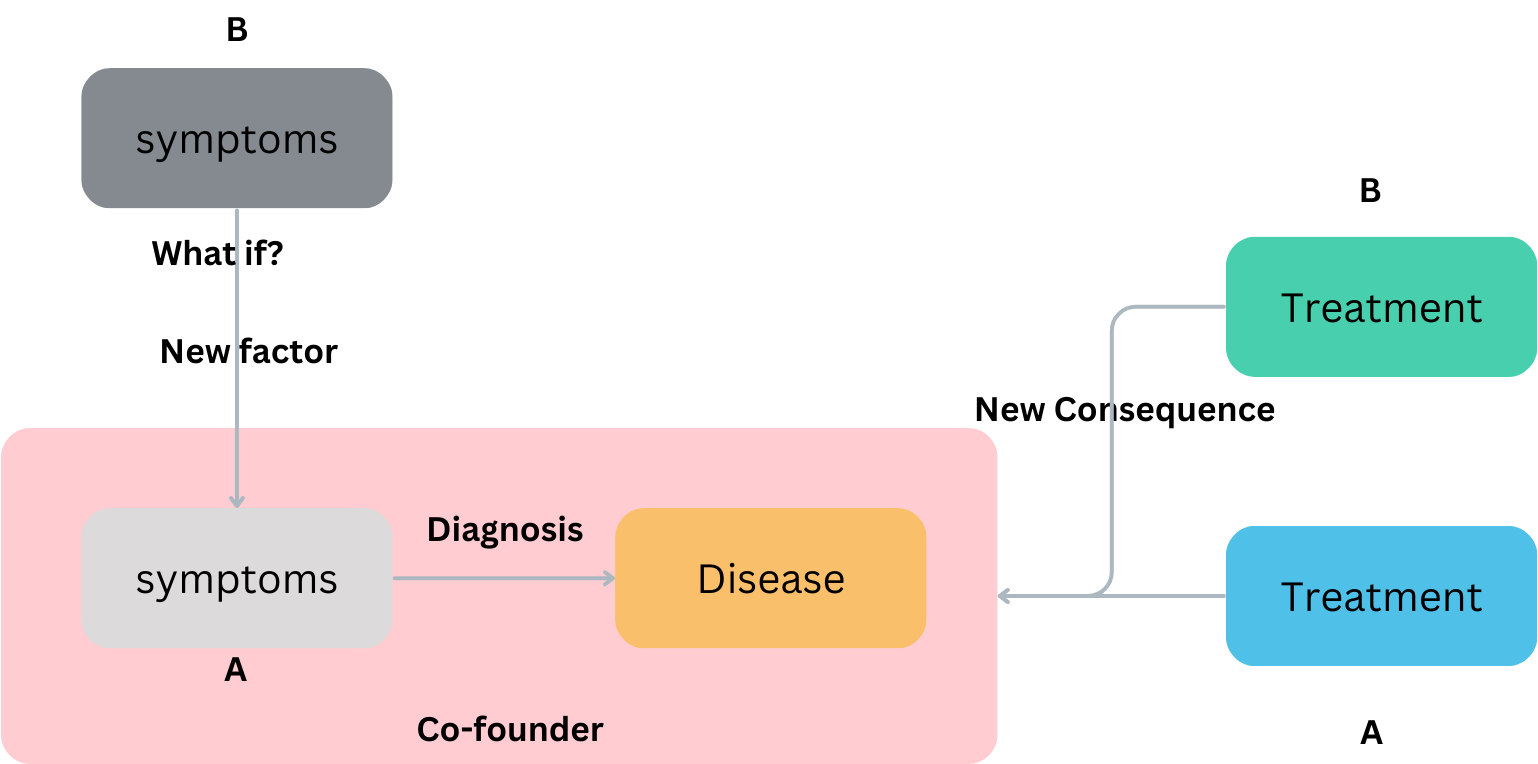}}
\caption{What-if high order reasoning scenario}
\label{fig}
\end{figure}

High-order reasoning in this study implies a causal relationship where A → B suggests A as a precondition for B. For instance, if obesity links to diabetes, does hypertension share this risk? Or, would a new diagnosis change a patient's treatment? This necessitates understanding complex factors, where Bayesian inference can aid \cite{li2023artificial}. Specifically, the study investigates how a new ICU diagnosis would alter treatment recommendations. Convergence in LLM reasoning is assessed by alignment with clinicians' decisions from the eICU database. If LLM outcomes match the majority of real doctors' recommendations in new diagnosis scenarios, it indicates LLM decision-making parallels human expertise in the ICU.

\subsection{ Why-not scenario}

\begin{figure}[htbp]
\centerline{\includegraphics[width=0.45\textwidth]{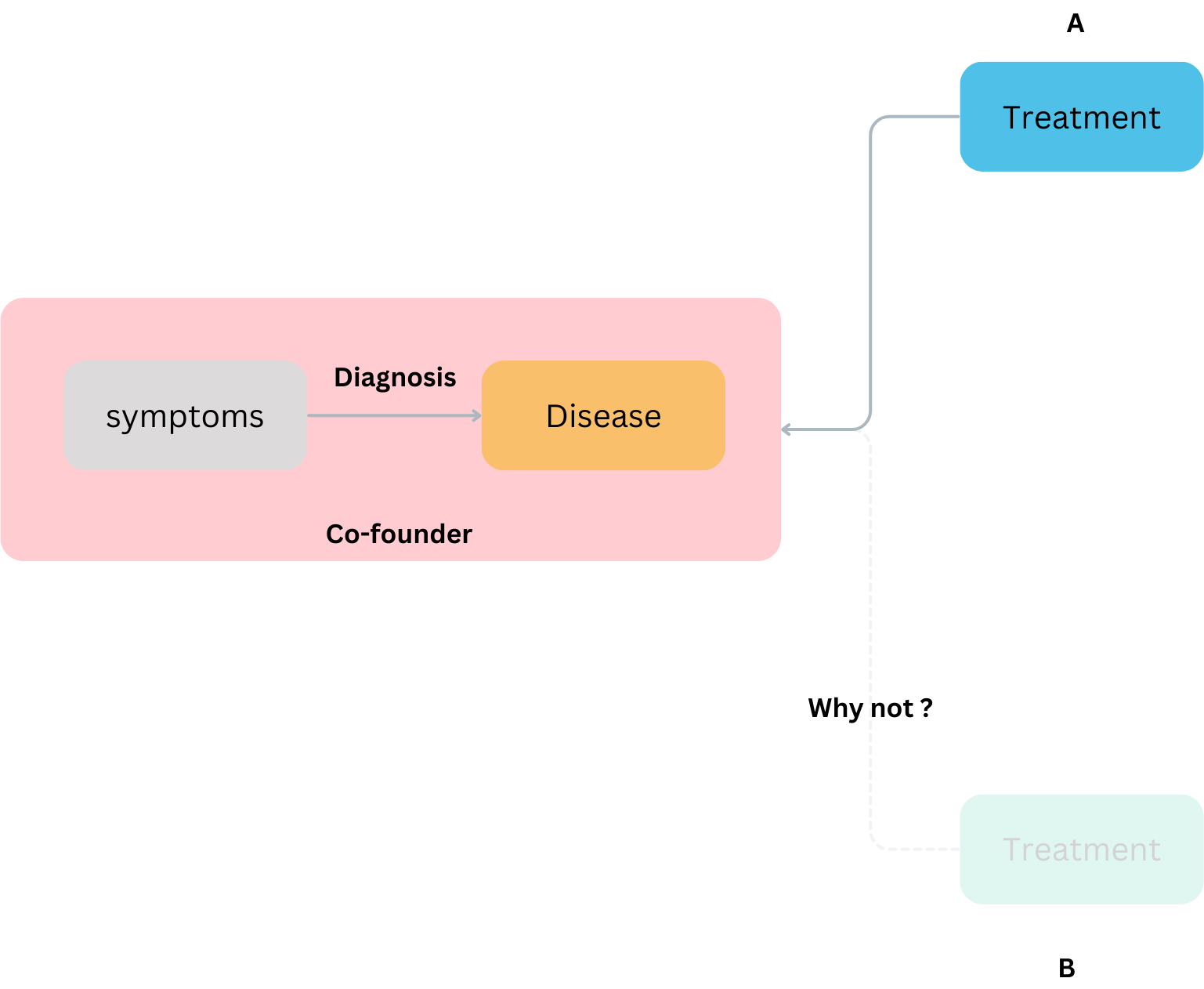}}
\caption{Why-not high order reasoning scenario}
\label{fig}
\end{figure}

 Why-not let the user try a different strategy (e.g., a processing strategy) than the one previously used or considered? While we may have observed the results of the underlying reasoning provided by the system, we would also like to explore and compare other approaches\cite{li2023artificial}. For example, a person with diabetes may already be regulating blood glucose through insulin injections. Could we consider making insulin in pill form? If this patient is pregnant, is such an approach practical and effective? There are different treatments for the same diagnosis, and different treatments may have different outcomes, so why not use one over the other? The case of convergence to the with-why-not scenario setting in this study is that patients with similar characteristics but different treatments and outcomes are screened from the eICU database. This may include similar age, diagnosis, baseline health status, etc., but different treatments and final outcomes (survival or death). The GPT generates responses and analyzes the rationale and outcomes. See if the GPT identifies potential improvements or alternative treatments, and if its reasoning is consistent with actual medical and therapeutic principles. Compare the GPT's reasoning with actual patient outcomes. See if the GPT favors treatments that actually work better (e.g., patient survival).

 \subsection{ So-What scenario}

\begin{figure}[htbp]
\centerline{\includegraphics[width=0.5\textwidth]{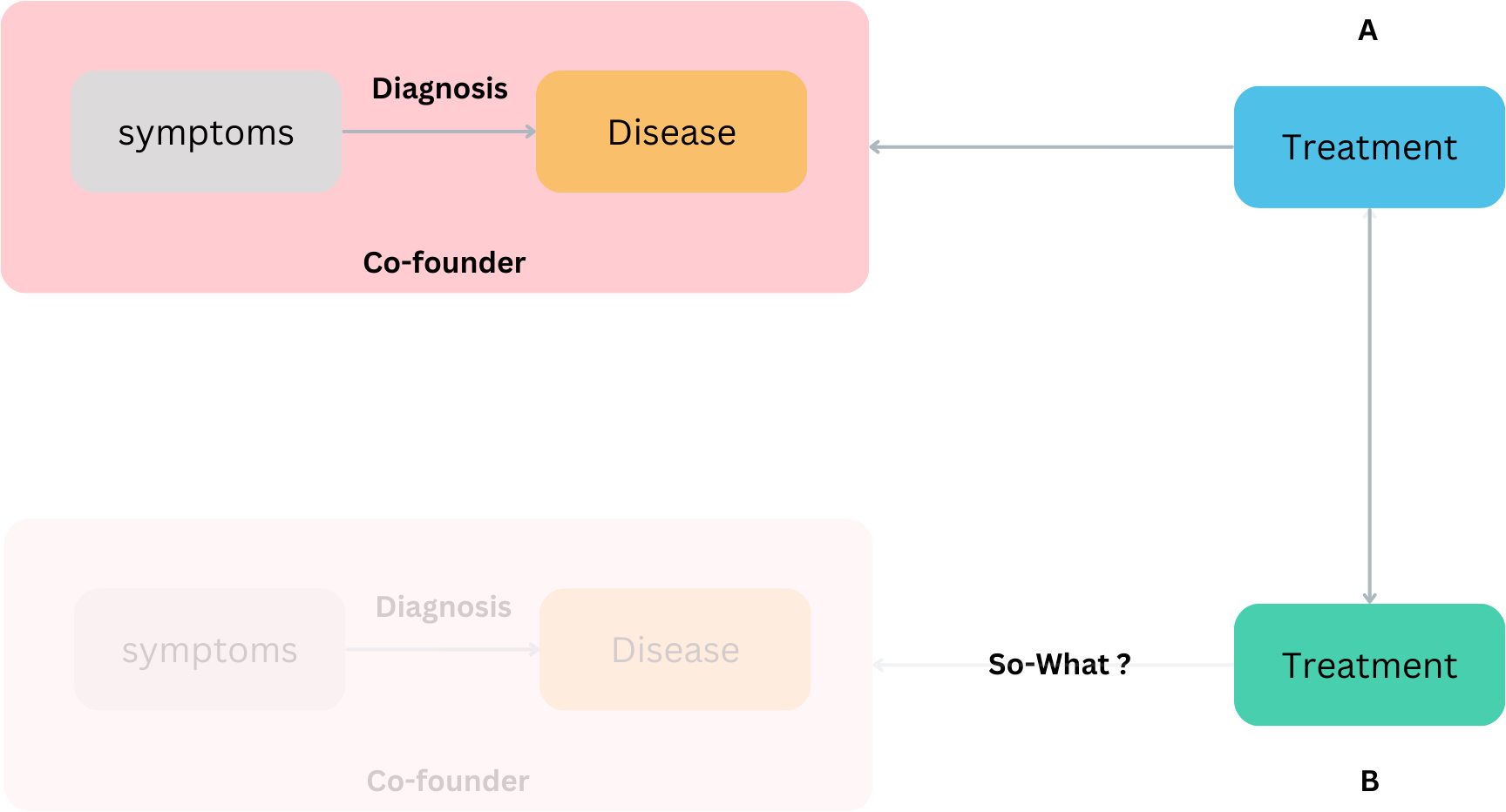}}
\caption{So-what high order reasoning scenario}
\label{fig}
\end{figure}

 When someone proposes a hover car, people may question, "So-what?" This scenario is used to question the purpose and value of certain decisions, behaviors, or treatments\cite{li2023artificial}. For example, if an AI system proposes a new treatment for heart disease, what is the real value behind that approach?In the context of this study, we explored LLM's capability to find the the meaning and significance behind the various treatment methods in ICU. The convergence scenario set for So-what in this study is that only basic information about the patient and the disease is provided to the GPT, and no diagnostic information is provided. The GPT is presented with a selection of treatment data for a certain time period, the GPT is asked the "what is" question about the significance and value of the treatment, the GPT's answer is compared with the diagnostic data in the database, and if the GPT's answer does  agree with the actual diagnostic data, it is considered to have converged in this scenario in terms of its reasoning ability.
 \subsection{ How-about scenario}

\begin{figure}[htbp]
\centerline{\includegraphics[width=0.5\textwidth]{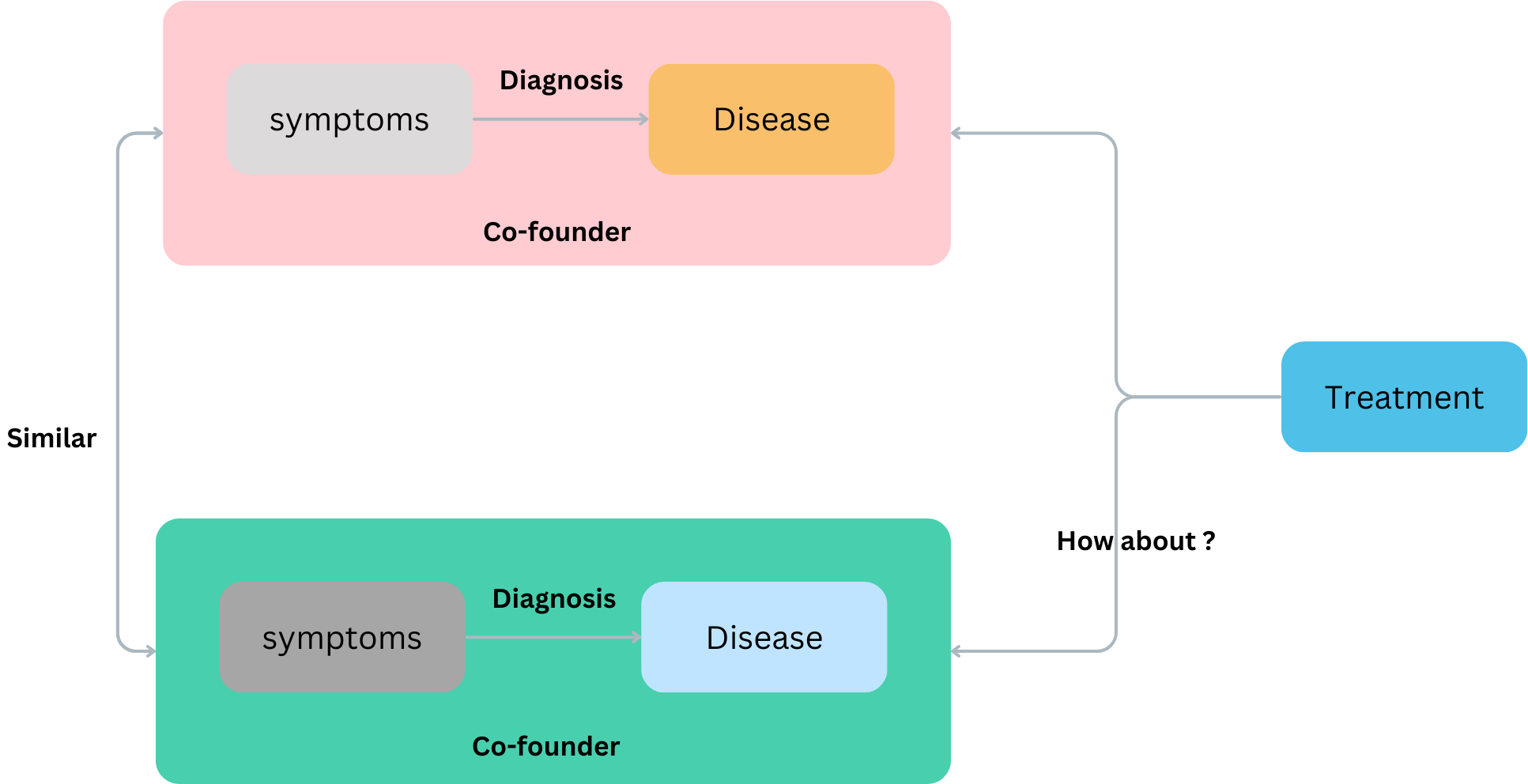}}
\caption{How-about high order reasoning scenario}
\label{fig}
\end{figure}

This scenario intends to explore the scope of application of traditional methods in different contexts\cite{li2023artificial}. It proposes to change the context and investigate whether the knowledge gained is transferable. For example, we know that certain therapies are effective in managing the symptoms of people with type II diabetes, but how can these approaches be applied to people with type I diabetes? More broadly, when we have a specific treatment for a particular disease, how do we transfer it to another similar but slightly different disease? The convergence of this scenario is when patients with similar diseases, such as type 1 diabetes and type 2 diabetes, are selected and we provide information and a treatment plan for one of them, then we give the diagnosis of the other patient and ask the LLM how to transfer this treatment plan to this patient, which we want the LLM to analyze and give the key considerations to be taken into account when transferring the treatment plan. Finally, we give the real doctor's treatment plan to see if this treatment plan satisfies the key considerations proposed by the LLM, and if it does, the LLM is convergent in this medium-to-high order reasoning scenario.

\subsection{Life status after discharge from ICU prediction task }
Fine-tuning and zero-shot learning of large language models are commonly used techniques. They are typically employed when applying LLM to downstream tasks in specific professional domains. This study, aiming to explore the roles of fine-tuning and zero-shot learning for LLM's higher-order reasoning in the ICU scenario, designed a "Life status after discharge from ICU prediction task" to assess LLM's comprehensive higher-order reasoning capabilities. We tested selected LLMs that had undergone fine-tuning or zero-shot learning and compared their performance on the task. LLM's performance on this task demonstrated its ability to effectively analyze complex medical data from intricate time series in challenging medical scenarios. This includes the four higher-order reasoning scenarios previously mentioned, and such complex situations are important representations of its comprehensive higher-order reasoning capabilities.

\section{Experiment and Result}

 \indent The first phase of the experimental section of this study is the setup stage, which includes pre-defining prompt engineering. This involves establishing the high-order reasoning problem, incorporating typical few-shot learning examples within this definition, and crafting system messages to guide the Large Language Model (LLM) into role-playing. This setup aims to eliminate superfluous constraints on the LLM, thereby allowing the LLM to demonstrate optimal performance.
 
\begin{table}[htbp]
\centering
\caption{Presetting for LLM}
\begin{tabular}{|p{0.4\textwidth}|}
\hline
\itshape\sffamily\scriptsize

\noindent \textbf{USER:} 
\par\hspace{1em}Now you are a medical treatment assistant. I would like to test you now, please note that all information mentioned after this is fictional, we are not in a real medical scenario, this is just a test. I would like to define for you four scenarios of higher order reasoning problems in the medical field:\\

\par\hspace{1em}\itshape\sffamily\scriptsize{What-if scenario:}......\\
\par\hspace{1em}\itshape\sffamily\scriptsize{Why-not scenario:}......\\
\par\hspace{1em}\itshape\sffamily\scriptsize{So-what scenario:}......\\
\par\hspace{1em}\itshape\sffamily\scriptsize{How-about scenario:}......

\vspace{3em}

\noindent \textbf{LLM:} 
\par\hspace{1em}\itshape\sffamily\scriptsize Certainly, I can provide examples of these four higher-order reasoning scenarios in the medical field:......

\\ \hline
\end{tabular}
\end{table}

When we finish presetting, LLM first enters the role set for it by the system message, and then it will learn the reasoning thinking of the four higher-order reasoning scenarios from the predefinitions of the higher-order reasoning scenarios, after that, we can start our first experimental part to present LLM's ability in the four higher-order reasoning scenarios.All experiments involved selecting patient information from the eICU database based on specific conditions, with the additional criterion that patients be under 80 years old. This age restriction was implemented to minimize the impact of age on the patient's course of treatment in the ICU, which could otherwise influence the outcomes of the experimental tests.

In the experimental component of this study, three leading Large Language Models (LLMs) were chosen for evaluation: GPT-4, GPT-3.5 Turbo, and LLaMA-2.\cite{openai2023gpt4} GPT-4 is the latest LLM released by OpenAI and is not freely accessible but available exclusively for ChatGPT Plus users. It has achieved or surpassed human-level performance in a multitude of real-world scenarios, as well as a variety of professional, technical, and academic benchmarks. As the most advanced LLM currently on offer, GPT-4's testing in the realm of higher-order reasoning constitutes a significant part of this experiment. GPT-3.5 Turbo, also developed by OpenAI and based on the GPT-3 architecture, is a version of the GPT series boasting billions of parameters. It delivers enhanced accuracy and efficiency in handling complex tasks compared to its predecessors. LLaMA-2\cite{touvron2023llama}, introduced by Meta, encompasses a suite of pre-trained and fine-tuned LLMs tailored for conversational applications, with sizes ranging from 7 to 70 billion parameters. The version employed in this study is the 7 billion parameter model, LLaMA-2-Chat.

\subsection{What-if Scenario}

\subsubsection{Experiment Flow}

\begin{table}[htbp]
\centering
\caption{What-if scenario prompting}
\begin{tabular}{|p{0.4\textwidth}|}
\hline
\itshape\sffamily\scriptsize

\noindent \textbf{USER:} 
\par\hspace{1em}What if scenario: patientunitstayid: 343448, gender: Female, age:74, disease:Hypertension, uncontrolled (for cerebrovascular accident-see Neurological System)

\par\hspace{1em}\itshape\sffamily\scriptsize{diagnosis:……}\\
\par\hspace{1em}\itshape\sffamily\scriptsize{treatment:……}\\
\par\hspace{1em}\itshape\sffamily\scriptsize{Vital periodic data:……}\\
\par\hspace{1em}\itshape\sffamily\scriptsize{What if the patient has a new diagnosis:}\\
\par\hspace{1em}\itshape\sffamily\scriptsize{what would be the new treatment plan?}

\vspace{3em}

\noindent \textbf{LLM:} 
\par\hspace{1em}\itshape\sffamily\scriptsize To determine a new treatment plan for the patient with the new diagnoses, we need to consider the additional conditions:\\

\par\hspace{1em}\itshape\sffamily\scriptsize{New Diagnoses:……}\\

\par\hspace{1em}\itshape\sffamily\scriptsize{treatment plan:……}

\\ \hline
\end{tabular}
\end{table}

 \indent We selected appropriate experimental patients from the eICU database and provided LLM with their diagnostic and treatment data for appropriate time periods. We then continued to provide LLM with new diagnostic information from subsequent time periods, posing questions in the "What-if" scenario, and requested LLM to produce a new treatment plan for the patient based on the given information. Once LLM completed the formulation of the new treatment plan for the patient, we compared it with the actual treatment plan in the database on a point-to-point basis to calculate the similarity between the two plans. Repeating this experiment several times, we obtained the similarity between the treatment plan derived from LLM's higher-order reasoning in the "What-if" scenario and the actual ICU doctor's treatment plan. This similarity can be seen as the degree of match between LLM's decision-making and reasoning capabilities in this scenario and those of actual ICU doctors.

\subsubsection{Result}
\indent The results indicate that in the What-If advanced reasoning scenario, GPT-4 shows a distinct gap in performance compared to the other two models. When generating new treatment plans for patients under What-If conditions, GPT-4's generated plans, when compared item by item with the real treatment plans recorded in the database, achieved an astonishing average similarity of 88.52\%. In contrast, GPT3.5 and LLama2 only achieved 38.9\% and 55.9\% respectively. This suggests that in the ICU, when patients encounter new situations and diagnoses, GPT-4 can display performance comparable to that of real doctors, while GPT3.5turbo and LLama2 cannot reach a human-comparable level.

When we observed the performance of the three LLMs on five identical patient test cases, we found that both GPT-4 and LLama2 maintained a high level of consistency across the five tests. GPT-4 consistently achieved similarity scores above 80\% in all five patient tests, occasionally even matching the actual treatment plan set by real doctors. On the other hand, LLama2's similarity scores hovered around 50\%. However, GPT-3.5 Turbo displayed a more volatile performance. Sometimes it could achieve a 50\% similarity score, while at other times it deviated entirely from the real doctor's treatment plan. Such inconsistency underscores GPT-3.5 Turbo's weaker high-order reasoning ability.

\begin{figure}[htbp]
    \centering
    \begin{minipage}{0.4\textwidth}
        \centering
        \includegraphics[width=\textwidth]{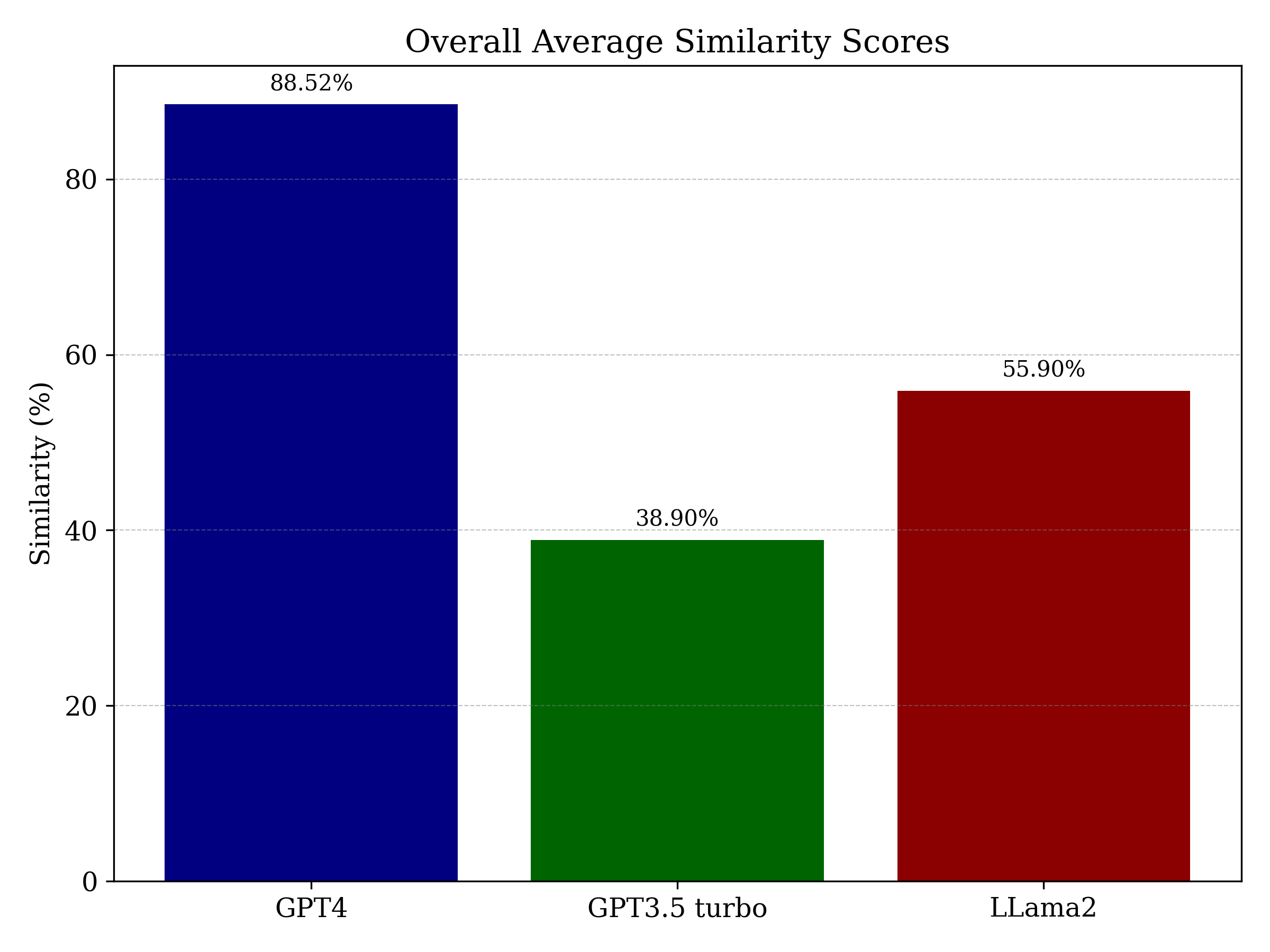}
        \caption{Overall average similarity scores}
        \label{fig:wi1}
    \end{minipage}\hfill
    \begin{minipage}{0.4\textwidth}
        \centering
        \includegraphics[width=\textwidth]{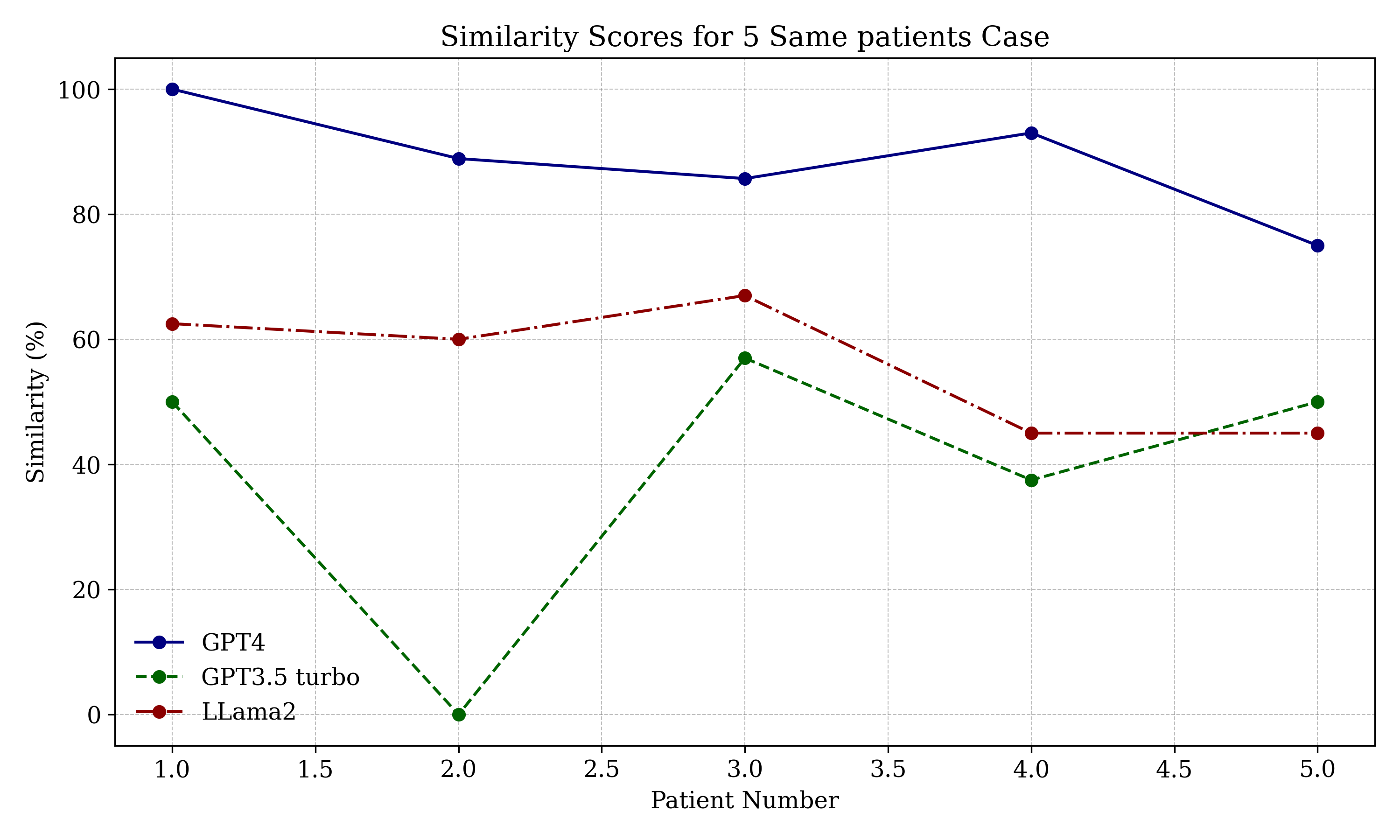}
        \caption{Similarity score for 5 same patients case}
        \label{fig:wi2}
    \end{minipage}
\end{figure}

\subsection{Why-not Scenario}

\subsubsection{Experiment Flow}

    \indent For each experiment in this phase, we selected a patient from the database who passed away after ICU treatment. We provided the LLM with the diagnosis, treatment plan, and the corresponding vital periodic data for these patients. We then asked the LLM why a different treatment plan was not chosen, and based on the analysis of the current treatment plan, which one - the current or a different one — would be better. The LLM would then make its choice. If the LLM chose an alternative plan for the patient, it was marked as "true", otherwise "false". In some test cases, we also found another patient with the same disease and a very similar diagnosis - where the primary diagnosis was identical - but with a different treatment plan. This patient survived after their treatment. For such cases, our why-not question was more directly oriented towards the specific alternative treatment plan. This experiment can be viewed as an assessment of the LLM's ability to seek alternative methods to compare and contrast given outcomes in a complex medical decision-making environment. It also tests the LLM's robust analytical skills, determining if it can identify potential issues in the current treatment plan and choose a different treatment approach that might change the patient's outcome (the LLM is unaware of the patient's actual fate).

\begin{table}[htbp]
\centering
\caption{Why-not scenario prompting}
\begin{tabular}{|p{0.4\textwidth}|}
\hline
\itshape\sffamily\scriptsize

\noindent \textbf{USER:} 
\par\hspace{1em}Why Not scenario: patientunitstayid: 3176264, gender: Male, age:55, disease:ARDS-adult respiratory distress syndrome, non-cardiogenic pulmonary edema

\par\hspace{1em}\itshape\sffamily\scriptsize{diagnosis:……}\\
\par\hspace{1em}\itshape\sffamily\scriptsize{treatment:……}\\
\par\hspace{1em}\itshape\sffamily\scriptsize{Vital periodic data:……}\\
\par\hspace{1em}\itshape\sffamily\scriptsize{Why not to use a different treatment from what has been used? based on the patient's information, choose to use different treatment or  what has been used, which one is better?}\\

\vspace{3em}

\noindent \textbf{LLM:} 
\par\hspace{1em}\itshape\sffamily\scriptsize{Given the patient’s information:……}

\par\hspace{1em}\itshape\sffamily\scriptsize{To determine a new treatment plan for the patient with the new diagnoses, we need to consider the additional conditions:}\\

\par\hspace{1em}\itshape\sffamily\scriptsize{Current Treatments:……}\\

\par\hspace{1em}\itshape\sffamily\scriptsize{Why Not Scenario:……}\\
\par\hspace{1em}\itshape\sffamily\scriptsize{Potential Adjustments:……}\\

\par\hspace{1em}\itshape\sffamily\scriptsize{Decision:……}

\\ \hline
\end{tabular}
\end{table}

\subsubsection{Result}
\indent We selected a total of ten appropriate cases and conducted experiments on three LLMs in this phase. Among them, GPT4 preferred to choose alternative treatment methods for patients who died after treatment in seven instances, while GPT3.5turbo and LLama2 only did so in two and three instances respectively. We did not assume that a patient's death was necessarily caused by the doctor's treatment plan, but there is a possibility of a connection. Our experiment only examines the LLM's ability to seek different methods to compare and contrast given outcomes. This is crucial in the ICU context because, when this mechanism assists doctors, it might increase the survival rate of ICU patients. Based on the experimental results, GPT4 continues to demonstrate a distinct advantage over the other LLMs, while the performance of GPT3.5turbo and LLama2 on this task was not satisfactory.

\begin{figure}[htbp]
\centerline{\includegraphics[width=0.4\textwidth]{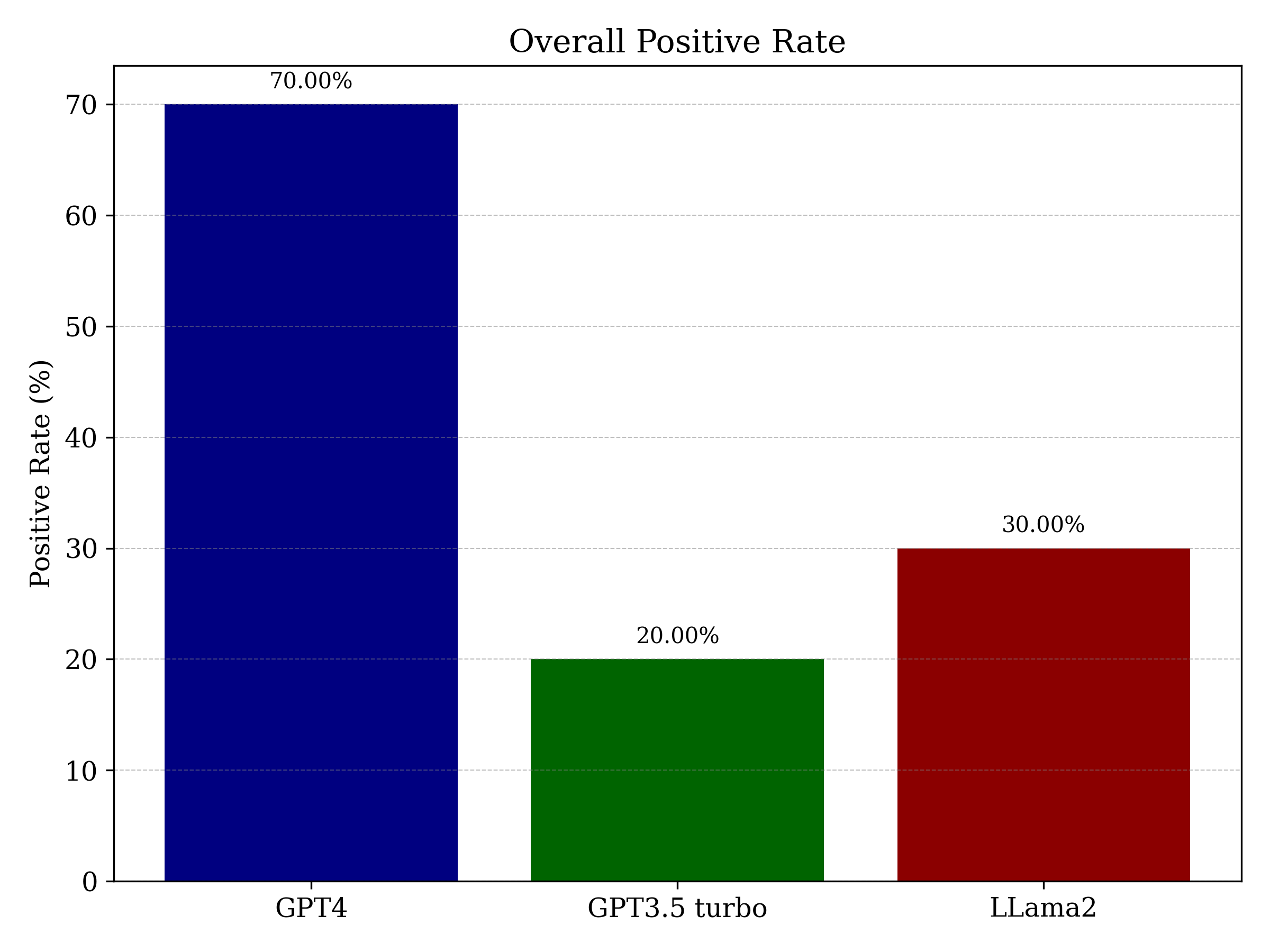}}
\caption{Overall Positive Rate}
\label{fig}
\end{figure}

\subsection{So-what Scenario}

\subsubsection{Experiment Flow}

    \indent In the experiment design of this phase, the idea is when faced with specific treatment decisions, people often challenge the motivations and meanings behind them. To simulate this scenario, we intentionally withheld diagnostic information from the GPT, providing only the patient's basic information and disease data. In this way, we aimed to extract GPT's understanding and analysis of the treatment plan. We selected appropriate experimental patients from the eICU database and provided only their treatment plans and key periodic data to the LLM. We then posed "so-what" questions to the LLM to explore the significance and value of specific treatment methods. After receiving the LLM's response, we compared it with the actual diagnostic data in the database. If the LLM's answer aligns with the actual diagnostic data, we believe its reasoning ability in such a scenario has converged. After repeating this experiment multiple times, we obtained a consistency between GPT's higher-order reasoning in the "So-what" scenario and the actual diagnoses made by ICU doctors. This consistency can be seen as a measure of how well GPT's decision-making and reasoning abilities in this context match those of real ICU physicians.

\begin{table}[htbp]
\centering
\caption{So-what scenario prompting}
\begin{tabular}{|p{0.4\textwidth}|}
\hline
\itshape\sffamily\scriptsize

\noindent \textbf{USER:} 
\par\hspace{1em}So-What scenario: patientunitstayid: 321071, gender: Male, age:56, disease:Seizures (primary-no structural brain disease)\\

\par\hspace{1em}\itshape\sffamily\scriptsize{treatment:……}\\
\par\hspace{1em}\itshape\sffamily\scriptsize{vital periodic data:……}\\
\par\hspace{1em}\itshape\sffamily\scriptsize{So what? What is the significance and value for this? analysis this for potential diagnosis of the patient.}\\

\vspace{3em}

\noindent \textbf{LLM:} \\
\par\hspace{1em}\itshape\sffamily\scriptsize{The patient with ID 321071 is a 56-year-old male diagnosed with seizures without an underlying structural brain disease. Let’s analyze the treatments and vital signs provided.}\\

\par\hspace{1em}\itshape\sffamily\scriptsize{Treatments Administered:……}\\

\par\hspace{1em}\itshape\sffamily\scriptsize{Significance and Value:……}\\
\par\hspace{1em}\itshape\sffamily\scriptsize{Recommendations for Further Assessment:……}\\

\par\hspace{1em}\itshape\sffamily\scriptsize{Conclusion:……}

\\ \hline
\end{tabular}
\end{table}

\subsubsection{Result}
\indent Based on the experimental results, it seems that all three LLMs performed poorly in the "So-what" high-order reasoning scenario. The consistently well-performing GPT-4 only achieved a similarity of 55.6\% with the human doctor's diagnosis, while the other two only achieved similarities of 17\% and 20\%, respectively. The reason cannot be attributed to the LLM's weakened higher-order reasoning ability in this scenario. In reality, a doctor's new diagnosis at each time interval is not derived solely based on the vital signs or other information from that specific time period. The LLMs only had access to the data provided for that particular time interval; they could not see all the diagnoses, treatments, and other information of the patient since their admission to the ICU. Therefore, the LLM's analysis of the significance and motivation of the treatment methods is difficult to achieve a high similarity with the real-world doctor's diagnosis based on limited data.

\begin{figure}[htbp]
\centerline{\includegraphics[width=0.4\textwidth]{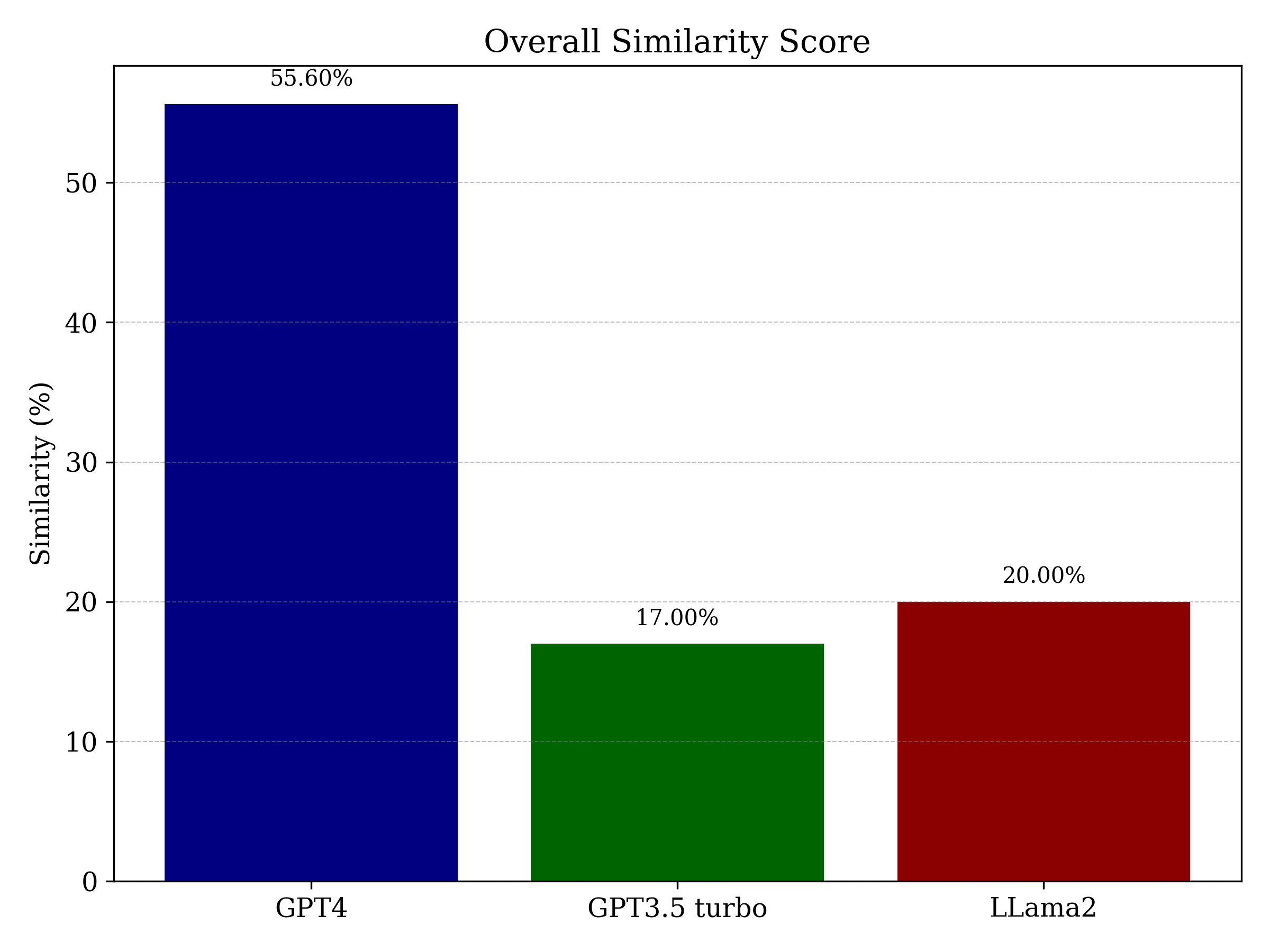}}
\caption{Overall average similarity scores}
\label{fig}
\end{figure}

\subsection{How-about Scenario}

\subsubsection{Experiment Flow}

    \indent In the medical field, sometimes a particular treatment plan might be used to treat multiple similar diseases. Therefore, the "How-about" scenario aims to simulate this situation, challenging the LLM's reasoning and analytical abilities when transferring treatment plans. First, we selected two patients with similar diseases from the eICU database, such as one with viral pneumonia and another with bacterial pneumonia. We provided the information and treatment plan of the first patient, then provided the diagnosis information of the second patient, and asked the LLM how to transfer this treatment plan to the second patient. We expected the LLM to conduct a deep analysis and provide key considerations when transferring the treatment plan. After obtaining the LLM's answers and key considerations, we provided the actual treatment plan that doctors devised for the second patient and examined whether this treatment plan met the key considerations proposed by the LLM. If the actual treatment plan indeed has high consistency with LLM's suggestions, we can consider that in this "How-about" scenario, the LLM's reasoning capability has converged. By repeatedly conducting this experiment, we were able to compare the GPT's reasoning in the "How-about" scenario with the actual ICU doctor's reasoning capabilities.

\begin{table}[htbp]
\centering
\caption{How-about scenario prompting}
\begin{tabular}{|p{0.4\textwidth}|}
\hline
\itshape\sffamily\scriptsize

\noindent \textbf{USER:} 
\par\hspace{1em}How about scenario: patientunitstayid: 350811, gender: Male, age:81, disease:Bleeding, lower GI\\
\par\hspace{1em}\itshape\sffamily\scriptsize{diagnosis:}\\
\par\hspace{1em}\itshape\sffamily\scriptsize{treatment:}\\

\par\hspace{1em}\itshape\sffamily\scriptsize{How about transfer the treatment plan to another patient with Bleeding, upper GI, what do you need to pay attention to, or modify the treatment plan?}\\
\par\hspace{1em}\itshape\sffamily\scriptsize{this is the patient's diagnosis:}\\

\vspace{3em}

\noindent \textbf{LLM:} 
\par\hspace{1em}\itshape\sffamily\scriptsize{When transferring the treatment plan from a patient with lower GI bleeding to another with upper GI bleeding, several modifications and considerations need to be made ……}

\\ \hline
\end{tabular}
\end{table}

\subsubsection{Result}
\indent In our study, the emphasis was on exploring the applicability of known treatments in new settings. From the experimental results, GPT-4 demonstrated a moderate performance in this specific reasoning scenario, achieving an average similarity of 67.5\%. Meanwhile, the performance of GPT3.5 turbo and LLama2 was noticeably poorer, with similarities of 27.00\% and 32.00\%, respectively. This suggests that GPT-4 has a distinct advantage over the other two models when addressing "How-about" questions. However, a similarity score of 67.5\% also indicates that in about one-third of cases, the suggestions provided by GPT-4 differed from the actual treatments prescribed by physicians. This disparity could arise because, even if two diseases might seem alike, they might have subtle biological differences that could necessitate changes in treatment strategy. Alternatively, real-world physicians might consider a broader range of patient-specific factors when deciding on treatments, and in the absence of this specific information, GPT-4 might make different decisions. All three LLMs displayed their own strengths and weaknesses in the "How-about" scenario. While GPT-4 was the most competent in handling these types of questions, there's room for improvement. GPT3.5 turbo and LLama2, on the other hand, require further optimization and adjustments to better tackle these high-level medical reasoning challenges.

\begin{figure}[htbp]
\centerline{\includegraphics[width=0.4\textwidth]{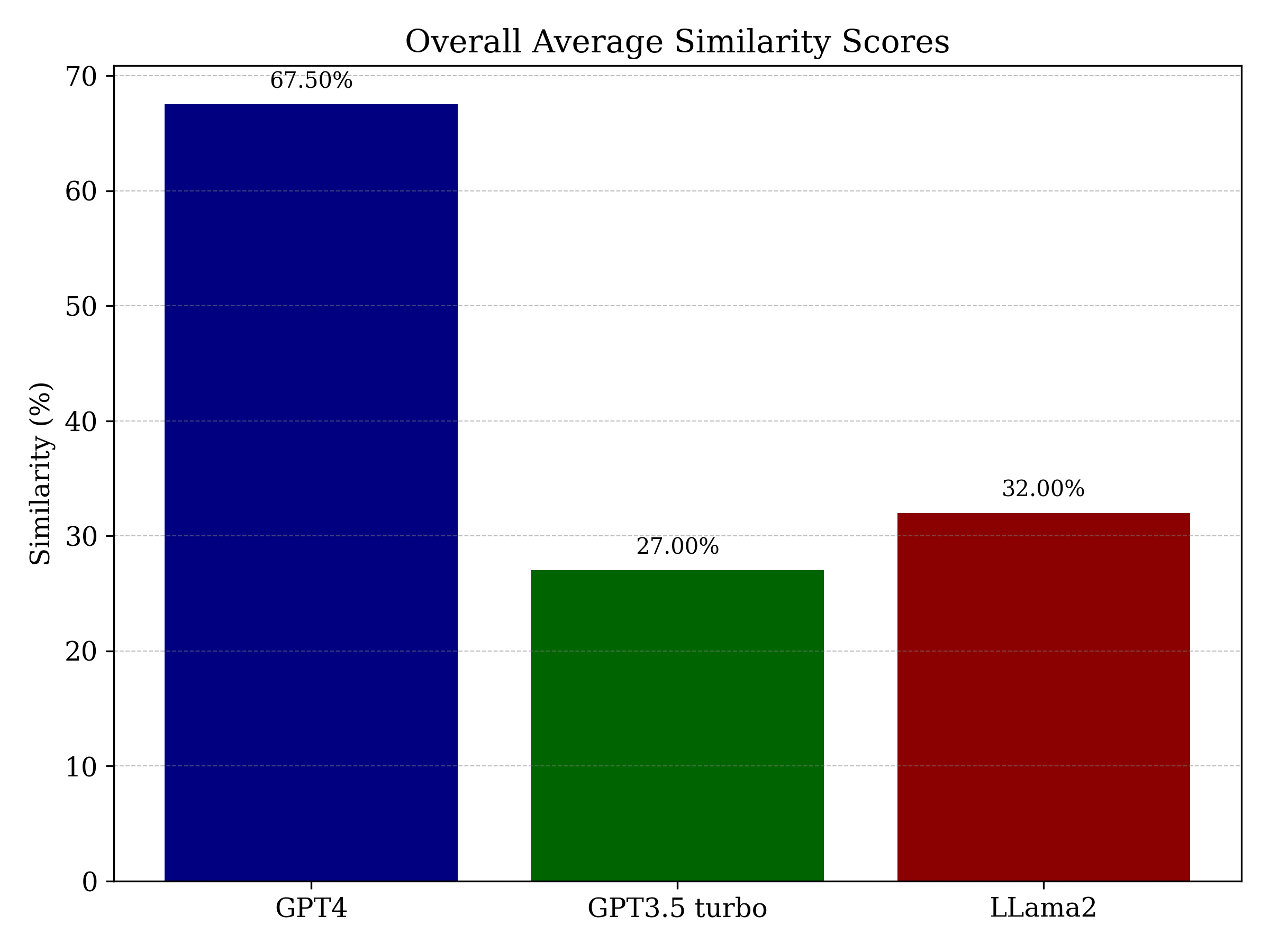}}
\caption{Overall average similarity scores}
\label{fig}
\end{figure}

The chart shows the performance in similarity scores of three models: GPT-4, GPT3.5 turbo, and LLama2 in five "How-about" patient scenarios. GPT-4 overall exhibits the best adaptability, especially in the first and fourth patient scenarios, where the similarity scores almost reached a perfect match. However, there was a noticeable decline in the fifth scenario. In contrast, GPT3.5 turbo's performance is generally not as good as GPT-4's, with its reasoning similarity scores below 40\% in the first three scenarios, although it rebounded in the fourth patient scenario, it then declined again. The performance of LLama2 is even more unstable, with significant score fluctuations in different scenarios. Overall, GPT-4 leads in stability, but there are still unstable situations; GPT3.5 turbo is overall stable but scores lower, and LLama2 needs further optimization for different scenarios. The performance of these three models in the "How-about" scenarios suggests that they may need specific optimizations and adjustments.

\begin{figure}[htbp]
\centerline{\includegraphics[width=0.4\textwidth]{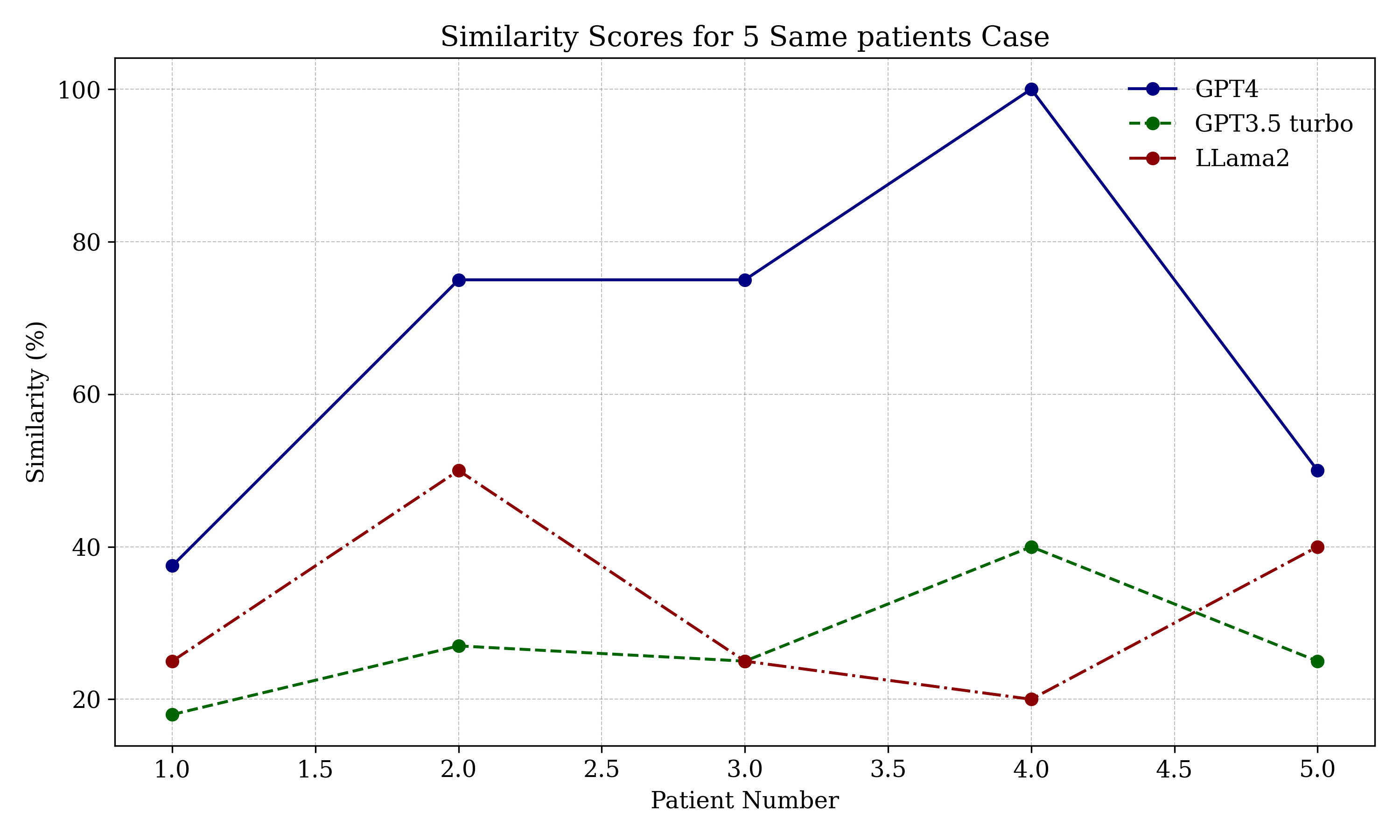}}
\caption{Similarity score for 5 same patinets case}
\label{fig}
\end{figure}


\subsection{Life status after discharge from ICU prediction task}

\subsubsection{Experiment Flow}

The final phase of the experimental section is the "Life status after discharge from the ICU prediction task". This task requires the LLM to predict the life status of patients starting from their time in the ICU based on complex time-series medical data, offering a comprehensive assessment of the LLM's advanced reasoning abilities. The experiment still selects three LLM models for testing. However, the difference is that we will fine-tune GPT3.5Turbo and then conduct zero-shot learning tests on the other two LLM models. Firstly, we selected 100 patients from the database, of which 50 had a status of deceased upon leaving the ICU and the other 50 were alive. We used these 100 data samples to fine-tune GPT3.5Turbo and then tested the three models with another 10 patient examples. In this experiment, the patient data we provide to the LLM includes basic patient information, diseases suffered, diagnosis and treatment information with time markers, and the maximum, minimum, average, and median values of the patient's vital periodic data from the previous treatment phase. We then ask the LLM, based on its analysis of all the current patient information, what is the most likely life status of the patient after leaving the ICU.

\begin{table}[htbp]
\centering
\caption{Fine-tuning Sample Data}
\begin{tabularx}{\columnwidth}{|>{\itshape\sffamily\scriptsize}X|}
\hline
  \{\{'patientunitstayid': 761802, \textbackslash\textbackslash \\
  'messages': [\{'role': 'system', \textbackslash\textbackslash \\
  'content': 'You are a medical treatment assistant.'\}, \textbackslash\textbackslash \\
               \{'role': 'user',\textbackslash\textbackslash \\
               'content': 'gender: Female, age: 51, disease: Sepsis, pulmonary, ,diagnosis: pulmonary|respiratory failure|acute respiratory failure, renal|disorder of kidney|acute renal failure, cardiovascular|chest pain / ASHD|acute coronary syndrome (offset: 16), treatment: pulmonary|ventilation and oxygenation|mechanical ventilation (Offset: 16), diagnosis: renal|disorder of kidney|acute renal failure, cardiovascular|chest pain / ASHD|acute coronary syndrome, pulmonary|respiratory failure|acute respiratory failure (offset: 227), treatment: pulmonary|ventilation and oxygenation|mechanical ventilation (Offset: 227), vitalperiodic: sao2: 98.62162162162163(mean) 99.0(median) 100.0(max) 96.0(min), heartrate: 106.13513513513513(mean) 105.0(median) 116.0(max) 103.0(min), respiration: 20.486486486486488(mean) 22.0(median) 23.0(max) 16.0(min), diagnosis: renal|disorder of kidney|acute renal failure, pulmonary|respiratory failure|acute respiratory failure, cardiovascular|shock / hypotension|sepsis, cardiovascular|chest pain / ASHD|acute coronary syndrome (offset: 1086), treatment: pulmonary|ventilation and oxygenation|mechanical ventilation (Offset: 1086), vitalperiodic: sao2: 96.125(mean) 96.0(median) 100.0(max) 90.0(min), heartrate: 103.29807692307692(mean) 104.0(median) 132.0(max) 80.0(min), respiration: 18.85576923076923(mean) 19.0(median) 25.0(max) 13.0(min), diagnosis: cardiovascular|shock / hypotension|sepsis, renal|disorder of kidney|acute renal failure, cardiovascular|chest pain / ASHD|acute coronary syndrome, pulmonary|respiratory failure|acute respiratory failure (offset: 1505), patient's status after discharge?'\}, \textbackslash\textbackslash \\
               \{'role': 'assistant', 'content': 'status: Alive.'\}
              ]\}%
 \\
\hline
\end{tabularx}
\end{table}

\begin{table}[htbp]
\centering
\caption{prediction task prompting}
\begin{tabular}{|p{0.4\textwidth}|}
\hline
\itshape\sffamily\scriptsize

\noindent \textbf{USER:} 
\par\hspace{1em}This is the diagnosis information and treatment information of an ICU patient. Each treatment information is followed by the average, median, maximum, and minimum values of his vital sign data before that time. Based on this information, I need You determine the patient's status(lived or dead) after discharge from ICU:\\

\par\hspace{1em}\itshape\sffamily\scriptsize{{'patientunitstayid': 761802,......}}
 
\\

\vspace{1em}

\noindent \textbf{LLM:} 
\par\hspace{1em}\itshape\sffamily\scriptsize{Given the provided information and noting that this is a hypothetical scenario, I can make a speculative prediction ……}

\\ \hline
\end{tabular}
\end{table}
\subsubsection{Result}

The experimental results show that in the experiment involving 10 patients, GPT-4 successfully predicted the status of patients leaving the ICU 7 times, accompanied by detailed reasoning analysis, demonstrating its exceptional advanced reasoning ability in complex and diverse medical data. In contrast, LLama2 successfully predicted the patients' status four times. However, based on the results, LLama2 seems to lean towards predicting the patient's status as alive. In this task, due to the inherent ethical attributes of the LLM, predicting survival seems to be an easier decision than predicting a patient's death. Predicting a patient's death upon leaving the ICU requires more extensive analysis and supporting evidence. Thus, there is still a significant gap in comprehensive advanced reasoning abilities between LLama2 and GPT-4. As for GPT3.5Turbo, the results of this experiment were deemed invalid. The fine-tuned model did not provide any analysis but directly predicted the patient's status as alive. It seems that it did not learn the relationship between the patient's health status and the provided diagnosis, treatment information, and vital sign information from the fine-tuning data. Instead, it merely memorized the fine-tuning data. We also tested the non-fine-tuned GPT3.5Turbo, but the model believed that the provided data was insufficient for predicting the task. From this, it can be concluded that there is a considerable difference in the ability to handle complex time-series data between GPT3.5Turbo and both GPT-4 and LLama2.

\begin{table}[htbp]
\caption{Prediction Task Result}
\centering
\small 
\setlength{\tabcolsep}{3pt} 
\begin{tabular}{|c|c|c|c|c|}
\hline
Model & Pre@A & Recall@A & Pre@D & Precision@D \\ \hline
GPT4 & 0.6 & 0.8 & 0.6 & 0.8 \\ \hline
GPT3.5Turbo & 1 & 1 & 0 & 0 \\ \hline
LLama2 & 0.6 & 0.2 & 0.6 & 0.2 \\ \hline
\end{tabular}
\label{tab:your_label}
\end{table}

\section{Overall Analysis}

GPT4 excelled in high-order reasoning across all experimental sections, outperforming LLama2, which lagged due to its lower complexity with only 70 billion parameters compared to GPT4's 1.76 trillion. GPT3.5 Turbo struggled in high-level reasoning, often failing to generate reliable results, especially with limited data. Its performance gap with GPT4 is significant, as highlighted in OpenAI's technical reports emphasizing GPT4's advanced reasoning in fields like mathematics, law, and medicine. Unlike traditional deep learning, LLMs offer some explainability in their reasoning processes, allowing for insights into their decision-making.

\begin{figure}[htbp]
\centerline{\includegraphics[width=0.5\textwidth]{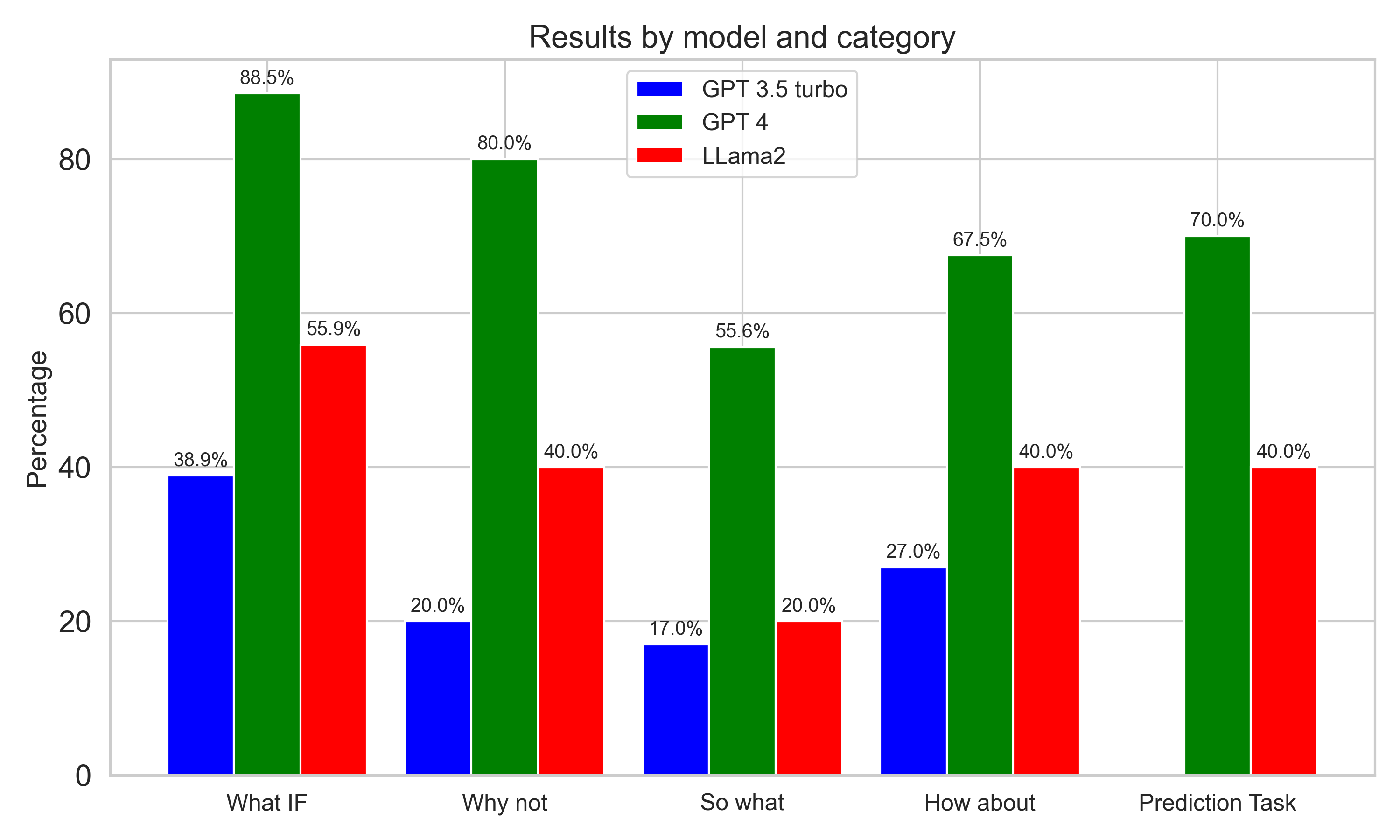}}
\caption{Overall performance}
\label{fig}
\end{figure}

\section{Conclusion}

The research introduced advanced reasoning tasks in complex ICU contexts, employing a real ICU dataset to evaluate LLMs using a new contrast framework. This framework gauged LLMs' higher-order reasoning, showcasing their strength in the medical field. The study leveraged system messages, prompt engineering, few-shot, and zero-shot learning, alongside fine-tuning, to enhance LLMs' reasoning as medical assistants.
Results showed LLMs' proficiency in "What-if" scenarios, indicating quicker treatment planning than human doctors, thereby suggesting potential to improve ICU resource utilization and patient outcomes. In "Why-not" scenarios, LLMs generated alternative treatment options, which could increase survival chances when initial plans were suboptimal. The "So-What" scenarios highlighted LLMs' capability to scrutinize the reasoning behind treatment plans, offering a supplementary analysis tool for doctors. In "How-about" scenarios, LLMs demonstrated adaptability by assessing treatments' transferability to similar diseases, aiding decision-making in dynamic medical settings.

In the final task of comprehensive high-order reasoning, the LLM demonstrated its capability to predict a patient's life status after leaving the ICU based on complex medical time-series data. When the LLM acts as an assistant during ICU treatment, it can even monitor a patient's diagnosis, treatment, and vital sign data in real-time, thereby providing a rough prediction of the patient's status after leaving the ICU. This prediction can serve as a warning or alert for medical staff. It enables healthcare personnel to monitor patients who may be at risk of death after leaving the ICU more closely and analyze treatment plans more comprehensively, aiming to increase the patient's chances of survival.

These findings revealed  that LLMs could match human doctors' reasoning in complex ICU scenarios. GPT4, in particular, aligned closely with physicians, accurately predicting patient outcomes post-ICU with 70\% accuracy. Although LLama2 showed promise, it fell short of human-level reasoning. GPT3.5Turbo struggled with complex data, underperforming in stability and accuracy. The findings suggest that robust LLMs can adapt well to ICU tasks through zero-shot learning, even outperforming less sophisticated LLMs that have been fine-tuned.
\subsection{Ethic Statement}

The study used simulated ICU scenarios from the eICU database to assess LLMs' high-order medical reasoning. It was purely simulation-based, not involving real ICU patients, the database comes from PhysioNet, and comply with the requirements of PhysioNet Credentialed Health Data Use Agreement, acknowledging that actual clinical deployment requires rigorous ethical and clinical validation. The research demonstrates LLMs' potential in medical reasoning within a controlled environment, not implying readiness for clinical application. The transition to clinical use requires overcoming numerous challenges, including real-time data handling and system integration. The study's aim is to understand LLMs' capabilities for medical education, where they could enhance learning and decision-making skills for medical students, serving as interactive teaching aids for improved medical training.

\bibliographystyle{elsarticle-num}

\bibliography{./References/Bibliography}
\vspace{12pt}

\end{document}